\documentclass[sigconf]{acmart}
\usepackage{graphicx}
\usepackage{amsmath} 
\usepackage{multirow}
\usepackage{booktabs}
\usepackage[table]{xcolor}
\usepackage{makecell}
\usepackage{enumitem}
\setitemize{noitemsep,topsep=0pt,parsep=0pt,partopsep=0pt}
\setlist{topsep=0pt, leftmargin=*}
\setcopyright{none}

\begin{document}

\title{GUI-AC: Enhancing Continual Learning in GUI Agents}

\author{Can Lin}
\affiliation{%
  \institution{Beijing University of Posts and Telecommunications}
  \city{Beijing}
  \country{China}
}
\email{lincan@bupt.edu.cn}

\author{Wenjun Yang}
\affiliation{%
  \institution{Alibaba Group}
  \city{Beijing}
  \country{China}
}
\email{lanshi.ywj@taobao.com}

\author{Tao Feng}
\affiliation{%
  \institution{Tsinghua University}
  \city{Beijing}
  \country{China}
}
\email{fengtao.hi@gmail.com}

\author{Hangjie Yuan}
\affiliation{%
  \institution{Zhejiang University}
  \city{Zhejiang}
  \country{China}
}
\email{hj.yuan@zju.edu.cn}

\author{Dan Zhang}
\affiliation{%
  \institution{National University of Singapore}
  \country{Singapore}
}
\email{zhangdan25@nus.edu.sg}

\author{Yifan Zhu}
\affiliation{%
  \institution{Beijing University of Posts and Telecommunications}
  \city{Beijing}
  \country{China}
}
\email{yifan\_zhu@bupt.edu.cn}

\author{Zhonghong Ou}
\affiliation{%
  \institution{Beijing University of Posts and Telecommunications}
  \city{Beijing}
  \country{China}
}
\email{zhonghong.ou@bupt.edu.cn}

\author{Wen Chen}
\affiliation{%
  \institution{Alibaba Group}
  \city{Beijing}
  \country{China}
}
\email{chenwen0718@foxmail.com}

\author{Jian Wu}
\affiliation{%
  \institution{Alibaba Group}
  \city{Beijing}
  \country{China}
}
\email{wujian.mailbox@gmail.com}

\renewcommand{\shortauthors}{Trovato et al.}

\begin{abstract}
    Graphical User Interfaces (GUIs) serve as the dominant medium for human–computer interaction, yet building GUI agents that generalize across the vast diversity of real-world interface environments, with the same flexibility and robustness that humans naturally exhibit, remains unsolved. Notably, GUI data are inherently non-stationary: the continual emergence of previously unseen interface instances (e.g., novel domains and resolutions) induces persistent distribution shifts, significantly impeding the continual learning of existing GUI agents. Reinforcement fine-tuning (RFT) has attracted considerable attention as a promising approach. Nevertheless, RFT exhibits pronounced instability in its grounding capability, where rollout outcomes are highly imbalanced and fixed clipping can suppress valuable exploration. More importantly, policy probability alone does not reveal the semantic role of a grounding prediction. To address these challenges, we propose GUI-\textcolor[HTML]{C71030}{A}\textcolor[HTML]{3170B5}{C}, a continual RFT framework that jointly characterizes grounding rollouts through advantage and policy probability. This yields a four-region view of grounding behavior and enables targeted optimization of Exploration and Overconfidence. GUI-AC further introduces grounding certainty to support two core mechanisms: (i) Adaptive \textbf{\textcolor[HTML]{C71030}{A}}dvantage, which strengthens the negative correction for overconfident grounding failures; and (ii) Dynamic \textbf{\textcolor[HTML]{3170B5}{C}}lipping, which relaxes the upper clipping bound for exploratory grounding rollouts. Extensive experiments show that GUI-AC consistently outperforms state-of-the-art baselines. Code is available at \url{https://anonymous.4open.science/r/GUI-AC-Code}.
\end{abstract}





\maketitle

\section{Introduction}
Graphical User Interfaces (GUIs) are the predominant means by which humans interact with digital devices~\cite{wang2024gui} and also serve as a key interaction layer for agents operating on Web applications. Unlike rigidly programmed automation, capable GUI agents are expected to adapt flexibly to previously unseen interface environments~\cite{tang2025survey,hong2024cogagent,guan2025kg}. This requirement is particularly important in Web scenarios, where agents may encounter substantial variations in application domains, interface layouts, and rendering resolutions. Such changes can significantly alter the spatial and visual grounding distribution~\cite{cheng2024seeclick,lin2025showui,guiaif}, making continual learning particularly challenging for GUI agents. This problem has been formalized as the Continual GUI Agents task~\cite{guiaif}. In this work, we focus on continual GUI grounding under cross-domain and cross-resolution shifts that are especially relevant to Web-oriented GUI agents.

\begin{figure*}[t]
  \centering
  \includegraphics[width=0.9\linewidth]{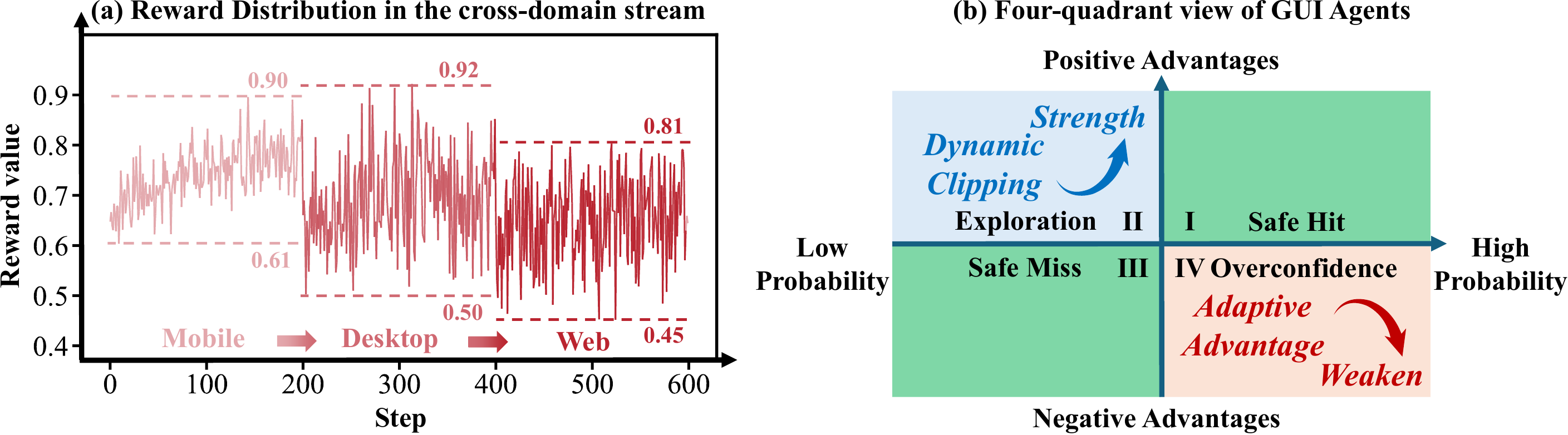}
  \caption{(a) The RFT-based method initially yields steady rewards. However, when transitioning to a new GUI domain, it shows sharp reward discontinuities and significant reward jumps at domain transitions. (b) Four-quadrant analysis of GUI-AC in the advantage-probability plane, where our method specifically targets the \textbf{Exploration} and \textbf{Overconfidence} regions.} 
  \label{fig:analysis}
\end{figure*}

GUI grounding is a fundamental capability of GUI agents, enabling them to map natural language instructions to pixel coordinates on interface elements~\cite{feng2022overcoming,du2025test,ye2025gui}. In continual learning scenarios, reinforcement fine-tuning (RFT) exhibits natural advantages~\cite{guiaif,lian2025ui}. GUI methods based on RFT leverage reward signals generated from online interactions, perform on-policy updates for self-correction, and thus achieve better grounding capability~\cite{liu2025visual,zhang2025reinforcement}. However, standard RFT typically assumes a relatively stable interface distribution. When the agent encounters unseen interface distributions across task boundaries, its grounding performance can degrade substantially~\cite{guiaif}, often accompanied by training instability~\cite{wang2025icpo,yang2025dcpo,chen2025empirical,yu2025dapo}. This limitation becomes particularly severe in continual GUI learning, where distribution shifts occur repeatedly over time. As shown in Figure~\ref{fig:analysis}(a), reward trajectories exhibit sharp discontinuities after domain transitions, revealing the vulnerability of standard RFT to continual interface shifts.

To identify the root causes of grounding degradation in continual GUI agents, we conduct a fine-grained analysis of policy dynamics. As shown in Figure~\ref{fig:analysis}(b), we map the agent’s states onto the four quadrants defined by advantage and probability. A complete four-quadrant analysis is deferred to Section~\ref{sec:rethinking}. The key challenges lie in two regions. Overconfidence corresponds to high-confidence actions inherited from previous interfaces. Under distribution shifts, these actions may no longer align with the new GUI geometry, resulting in inconsistent rewards and gradients across rollouts that amplify training noise and undermine localization stability. Exploration represents rare but correct actions under a novel interface and provides crucial gradient signals for adapting to the new distribution. However, the fixed clipping bound restricts the increase in their probabilities, suppressing effective exploration. This asymmetric optimization under distribution shift can rapidly reduce policy entropy and lead to premature policy collapse.

To address these challenges, we propose GUI-\textcolor[HTML]{C71030}{A}\textcolor[HTML]{3170B5}{C}, a method that enhances the continual learning capability of GUI agents. We introduce grounding certainty as an observable proxy for grounding capability within each sample group. Guided by the grounding certainty, GUI-AC jointly calibrates two complementary mechanisms: (i) Adaptive \textbf{\textcolor[HTML]{C71030}{A}}dvantage, which scales the normalized group advantage and calibrates noisy instances to prevent policy overconfidence; and (ii) Dynamic \textbf{\textcolor[HTML]{3170B5}{C}}lipping, which relaxes the clipping bound to selectively encourage the exploration range. These two mechanisms enable the policy to remain stable on previously seen tasks while actively exploring unseen interfaces. In summary, the main contributions of our paper are:
\begin{itemize}
    \item We revisit grounding instability in continual GUI agents and introduce a four-quadrant analysis to characterize the key failure modes of RFT under continual shifts.
    \item We propose GUI-AC, a novel method that enhances the continual learning capability of GUI agents through two core mechanisms: Adaptive Advantage and Dynamic Clipping.
    \item GUI-AC achieves state-of-the-art results on ScreenSpot V1, V2, and Pro benchmarks, delivering robust improvements in training stability and continual generalization.
\end{itemize}

\section{Related Works}

\subsection{GUI Agents}
GUI agents aim to understand natural-language instructions and execute corresponding low-level interactions to accomplish user tasks on digital interfaces~\cite{tang2025survey,hong2024cogagent,cheng2024seeclick,gou2024navigating,zhang2024large,sun2025genesis,yang2025aria,gao2025uishift}. Existing methods for GUI agents and grounding can be broadly categorized into two paradigms. (1) Expert design-driven workflow paradigm. This line of work typically uses proprietary multimodal models as high-level planners, and combines them with manually engineered planners~\cite{wang2024mobile,zhang2025ufo} or grounders~\cite{lin2025showui,gou2024navigating,liu2024autoglm,wu2024atlas} to form a controllable pipeline. Grounding is commonly achieved either by leveraging structural signals~\cite{zhang2025appagent} or by using vision-based tooling~\cite{kirillov2023segment,lu2024omniparser}. Workflow approaches are easy to deploy and interpret, yet often rely on external modules and hand-crafted rules, which can reduce robustness under domain shift and UI drift~\cite{qin2025ui}. (2) Data-driven training paradigm. Another line of work improves GUI perception and reasoning by scaling task-specific data or architectures~\cite{hong2024cogagent,qin2025ui}. Despite strong performance, these methods can still struggle to generalize to previously unseen interface styles~\cite{luo2025gui}.

\subsection{Continual Post-training for GUI Agents}
Adapting a base model to GUI grounding tasks generally relies on two post-training paradigms: Supervised Fine-Tuning (SFT) and Reinforcement Fine-Tuning (RFT). SFT can overfit the current label distribution and suffer from catastrophic forgetting under distribution shifts~\cite{guo2025continuallearninggenerativeai,liu2025llava,shenfeld2025rl}. By contrast, RFT, particularly with on-policy updates, is better suited to continual adaptation, but introduces optimization challenges when interface distributions evolve~\cite{liu2025visual,zhang2025reinforcement,jin2025rl}. Recent GUI agents have incorporated verifiable rewards into RFT, extending the R1 paradigm~\cite{guo2025deepseek} to GUI interaction~\cite{luo2025gui,lu2025ui,liu2025infigui,zhou2025gui}. GUI-G$^2$~\cite{tang2025gui} further improves grounding supervision through geometry-aware reward modeling, while GUI-AiF~\cite{guiaif} explicitly studies continual adaptation across changing GUI distributions. However, these methods mainly improve the reward signal or training objective without explicitly distinguishing the optimization roles of different grounding rollouts under distribution shifts. This motivates an optimization mechanism that jointly considers grounding outcomes and policy probability.

\section{Method}

\subsection{Problem Formulation}

The task of GUI grounding requires an agent to map a natural-language instruction and an interface to the pixel coordinates of the corresponding interactive element. Given an interface and an instruction, the model predicts a bounding box $\mathbf{b}^{p}=[x_{1}^{p},y_{1}^{p},x_{2}^{p},y_{2}^{p}]$, with $(x^{p}_{1},y^{p}_{1})$ and $(x^{p}_{2},y^{p}_{2})$ denoting the top-left and bottom-right corners. The ground-truth box is $\mathbf{b}^{gt}=[x^{gt}_{1},y^{gt}_{1},x^{gt}_{2},y^{gt}_{2}]$. 
We treat a prediction as correct if its center point 
$\mathbf{c}_{p}=\big(\tfrac{x^{p}_{1}+x^{p}_{2}}{2},\tfrac{y^{p}_{1}+y^{p}_{2}}{2}\big)$ 
lies within $\mathbf{b}^{gt}$. We focus on the Continual GUI Agents setting~\cite{guiaif}, which is modeled as tasks that arrive sequentially, including a cross-domain sequence $\mathcal{T}_{D}=\{ t_{d1}, t_{d2}, \ldots,  t_{dn}\}$ and a cross-resolution sequence $\mathcal{T}_{R}=\{t_{r1}, t_{r2}, \ldots, t_{rn}\}$. The objective is to learn a policy that continually adapts to newly arriving tasks while preserving grounding performance on previously encountered tasks. 

\begin{figure*}[h]
  \centering
  \includegraphics[width=\linewidth]{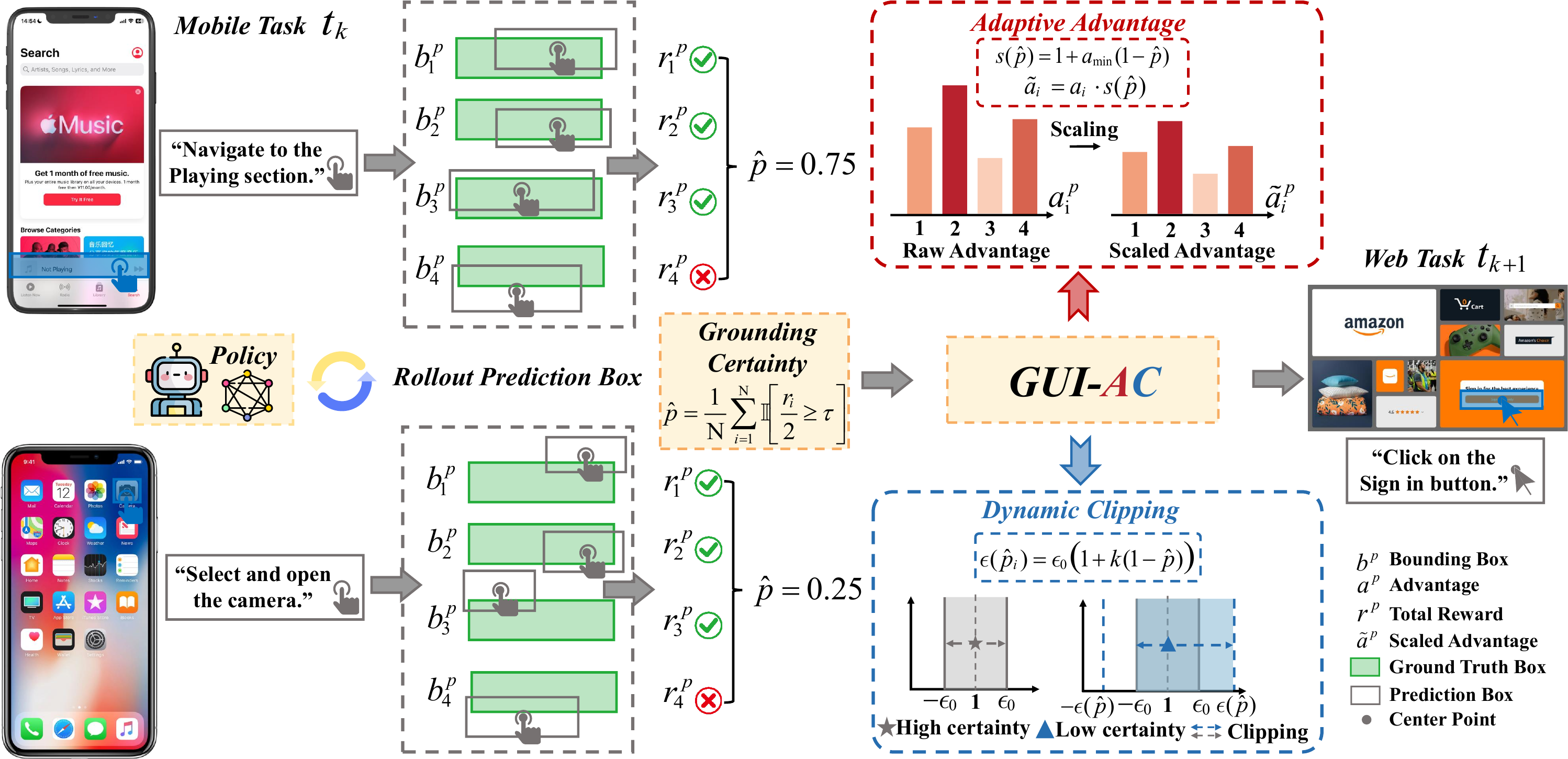}
  \caption{Illustration of GUI-\textcolor[HTML]{C71030}{A}\textcolor[HTML]{3170B5}{C}. \emph{Grounding Certainty} controls the strength of two complementary mechanisms: \emph{Adaptive \textbf{\textcolor[HTML]{C71030}{A}}dvantage}, which rescales advantages to calibrate overconfident rollouts, and \emph{Dynamic \textbf{\textcolor[HTML]{3170B5}{C}}lipping}, which adjusts clipping bounds to balance exploration.} 
  \label{fig:Framework}
\end{figure*}

\subsection{Rethinking Continual GUI Agents in RFT}
\label{sec:rethinking}


RFT-based methods are appealing for continual GUI agents due to online interaction and on-policy self-correction. However, under domain or resolution shifts, the same policy update can have qualitatively different effects on different sampled actions. Importantly, policy probability alone does not reveal the semantic role of a grounding prediction. We therefore rethink continual GUI agents through the joint lens of advantage and policy probability.

\textbf{Four-Quadrant View of Continual GUI Agents.} 
For each training instance $i$, we sample $N$ rollouts from $\pi_\theta$, obtaining rewards $\{r_{i,1},\ldots,r_{i,N}\}$. The reward of rollout $g$ is defined as $ r_{i,g}.$ The detailed reward definitions are provided in Appendix~\ref{app:reward}. Following GRPO, we compute the group-wise mean and standard deviation:
\begin{equation} 
\mu_i = \frac{1}{N}\sum_{g=1}^{N} r_{i,g}, \qquad \sigma_i = \sqrt{ \frac{1}{N}\sum_{g=1}^{N}(r_{i,g}-\mu_i)^2 }.
\end{equation}
To avoid division by zero when the rewards within a group are all equal, we define the normalized advantage with a small numerical stabilizer $\varepsilon > 0$:
\begin{equation} 
A_{i,g} = \frac{r_{i,g}-\mu_i} {\sigma_i+\varepsilon}, 
\end{equation}
We further characterize each rollout by its length-normalized probability under the rollout policy. Let $u_i=(s_i,x_i)$ denote the input context, where $s_i$ is the interface observation and $x_i$ is the instruction, and let $c_{i,g,t}=(u_i,y_{i,g,1:t-1})$ denote the decoding context for the $t$-th token. For rollout $g$ with $T_{i,g}$ generated tokens, we define \begin{equation} q_{i,g} = \exp\left( \frac{1}{T_{i,g}} \sum_{t=1}^{T_{i,g}} \log \pi_{\mathrm{old}} \left( y_{i,g,t}\mid c_{i,g,t} \right) \right). \end{equation} This quantity corresponds to the geometric mean of the token-level probabilities assigned by the rollout policy and satisfies $q_{i,g}\in(0,1]$. We compute 
\begin{equation} 
\bar q_i = \frac{1}{N} \sum_{g=1}^{N}q_{i,g}. 
\end{equation}
For a sampled action, its optimization effect is jointly determined by two factors: whether the action receives a positive or negative advantage, and whether the current policy assigns it a high or low probability. This yields four representative regions in the advantage-probability plane.

\begin{itemize}
\item \textbf{I: Safe Hit} ($A_{i,g}>0$, $q_{i,g}\ge\bar q_i$). Actions with high probability and positive advantage. These actions correspond to interaction patterns that the policy has already mastered and are often inherited from seen interfaces.

\item \textbf{II: Exploration} ($A_{i,g}>0$, $q_{i,g}<\bar q_i$). Actions with low probability and positive advantage. These actions are rare but valuable. Under a shifted GUI distribution, only a few sampled actions accidentally hit the correct element. They provide the most important corrective signal for adapting to the new interface.

\item \textbf{III: Safe Miss}($A_{i,g}<0$, $q_{i,g}<\bar q_i$). Actions with low probability and negative advantage. These actions are incorrect but usually less harmful to continual adaptation, since the policy already assigns them low probability.

\item \textbf{IV: Overconfidence}($A_{i,g}<0$, $q_{i,g}\ge\bar q_i$). Actions with high probability and negative advantage. These actions are the most problematic under distribution shift. The policy confidently predicts coordinates based on an outdated interface layout, but these predictions no longer match the current GUI geometry.
\end{itemize}

This quadrant view reveals two coupled challenges. (i) Which advantages should be trusted when high-probability actions become overconfident? (ii) How can rare low-probability correct actions be explored before they disappear from future rollouts? 

\textbf{Unreliable Advantage Estimation.}
When the distribution shifts, the rollout outcomes within the same group often become highly imbalanced. Group-based advantage estimation is easily corrupted by noisy high-confidence failures, amplifying optimization variance. This effect is especially severe for Overconfidence. Generic RFT stabilization mechanisms do not explicitly distinguish the semantic roles of grounding rollouts under interface shifts. In particular, probability- or confidence-based optimization signals alone cannot determine whether a sampled spatial prediction represents useful exploration or an erroneous grounding pattern. 

\textbf{Fixed Clipping Suppresses Exploration.}
When the distribution shifts, only a very small number of samples within a rollout group typically hit the correct element by chance and receive high rewards. Specifically, the most valuable signals come from Exploration. However, the fixed clipping mechanism commonly used in standard RFT constrains the magnitude of probability-ratio changes between the new and old policies. For correct actions with initially low probability, their probability increase is tightly restricted, making it difficult for them to be sufficiently amplified within one or only a few updates. Over time, the policy becomes increasingly concentrated on a small set of already high-probability behaviors. This causes the policy to become prematurely deterministic.

\subsection{GUI-AC Framework}

As illustrated in Figure~\ref{fig:Framework}, GUI-AC introduces
Grounding Certainty $\hat p_i$ to modulate two region-specific
mechanisms:
(i) \textit{Adaptive Advantage}, which strengthens the negative correction
for overconfident grounding failures; and
(ii) \textit{Dynamic Clipping}, which relaxes the upper clipping
bound for exploratory grounding rollouts.

\textbf{Measuring Grounding Capability via Certainty.} 
We introduce Grounding Certainty as an observable proxy for grounding capability. Since the rollout-specific grounding reward satisfies $r_{i,g}\in[0,2]$, we normalize it to $[0,1]$ by dividing by $2$. Grounding Certainty is then defined as \begin{equation} \hat p_i = \frac{1}{N} \sum_{g=1}^{N} \mathbb{I}\!\left[ \frac{r_{i,g}}{2}\ge\tau \right], \end{equation}
where $\tau\in[0,1]$ is a predefined grounding threshold. A high $\hat p_i$ indicates stronger group-level grounding capability, whereas a low $\hat p_i$ indicates greater mismatch with the current interface distribution. Unlike the within-group signals $A_{i,g}$ and $q_{i,g}$, which determine the optimization region of each rollout, $\hat p_i$ determines only the strength of the corresponding region-specific correction.

\textbf{Adaptive Advantage.} 
We first define the scaling factor:
\begin{equation} 
w_i = 1 + a_{\min}(1-\hat p_i). 
\end{equation} 
where $a_{\min}\in(0,1)$ controls the strength of advantage amplification.. We define a rollout-specific advantage: 
\begin{equation} \widetilde{A}_{i,g} = 
\begin{cases} w_i A_{i,g}, & A_{i,g}<0,\; q_{i,g}\ge \bar q_i, \\[2pt] A_{i,g}, & \text{otherwise}. 
\end{cases} 
\end{equation} 
For overconfident rollouts, when grounding certainty is low, unreliable high-confidence failures are more strongly suppressed.

\textbf{Dynamic Clipping.}
We define token-level probability ratio as:
\begin{equation}
\rho_{i,g,t}
=
\frac{
\pi_{\theta}(y_{i,g,t}\mid c_{i,g,t})
}{
\pi_{\mathrm{old}}(y_{i,g,t}\mid c_{i,g,t})
}.
\end{equation}
We first define the clipping margin as:
\begin{equation}
\epsilon_i = \epsilon_0 \big(1 + k(1 - \hat p_i)\big),
\end{equation}
where $\epsilon_0>0$ is the base clipping margin and $k\ge0$ is the
trust-region expansion factor. We define a rollout-specific upper bound: 
\begin{equation} U_{i,g} = 
\begin{cases} 1 + \epsilon_i, & A_{i,g}>0,\; q_{i,g}<\bar q_i, \\[2pt] 1+\epsilon_0, & \text{otherwise}. 
\end{cases} 
\end{equation}
We keep the standard lower clipping bound $L=1-\epsilon_0$, while selectively expanding the upper bound for exploratory rollouts. The clipped ratio is then defined as:
\begin{equation}
\mathrm{clip}_{i,g}(\rho_{i,g,t})
=
\min\big\{
\max(\rho_{i,g,t},L),
U_{i,g}
\big\}.
\end{equation}
For exploratory rollouts, lower grounding certainty produces a larger upper clipping bound, allowing their probabilities to increase more freely and thereby preserving exploration under distribution shifts.

\textbf{Reinforcement Fine-tuning with GUI-AC.}
The resulting clipped surrogate objective is:
\begin{equation} \ell_{i,g,t}(\theta) = \min\Big( \rho_{i,g,t}\widetilde{A}_{i,g}, \; \mathrm{clip}_{i,g}(\rho_{i,g,t}) \widetilde{A}_{i,g} \Big). 
\end{equation} 
The final objective is \begin{equation} \mathcal{J}(\theta) = \mathbb{E}_{i,g,t} \left[ \ell_{i,g,t}(\theta) - \beta D_{\mathrm{KL}} \left( \pi_{\theta}(\cdot\mid c_{i,g,t}) \Vert \pi_{\mathrm{ref}}(\cdot\mid c_{i,g,t}) \right) \right], \end{equation} where $\beta\ge0$ controls the strength of KL regularization.

\begin{table*}[h]
    \centering
    \caption{Continual domain performance (\%) of the proposed GUI-AC method on the ScreenSpot-V1 and ScreenSpot-V2 benchmark. The M. D. and W. denote the Mobile, Desktop and Web domain tasks, respectively. The \textcolor{gray}{upper bound} corresponds to the state-of-the-art performance achieved by RFT-based methods for GUI agents and is provided as a reference.
    }
    \label{screenspot12}
    \resizebox{\textwidth}{!}{ 
    
    \begin{tabular}{@{}l l cc cc cc c cc cc cc c@{}} 
        \toprule
        
        \multirow{3.5}{*}{\textbf{Method}} & \multirow{3.5}{*} & \multicolumn{7}{c}{\textbf{SSv1 Accuracy (\%)}} & \multicolumn{7}{c}{\textbf{SSv2 Accuracy (\%)}} \\
        
        \cmidrule(lr){3-9} \cmidrule(lr){10-16}
         & & \multicolumn{2}{c}{Mobile} & \multicolumn{2}{c}{Desktop} & \multicolumn{2}{c}{Web} & \multirow{2.5}{*}{\textbf{Avg.}} & \multicolumn{2}{c}{Mobile} & \multicolumn{2}{c}{Desktop} & \multicolumn{2}{c}{Web} & \multirow{2.5}{*}{\textbf{Avg.}} \\
        
        \cmidrule(lr){3-4} \cmidrule(lr){5-6} \cmidrule(lr){7-8} \cmidrule(lr){10-11} \cmidrule(lr){12-13} \cmidrule(lr){14-15}
         & & Text & Icon & Text & Icon & Text & Icon & & Text & Icon & Text & Icon & Text & Icon & \\
        
        \midrule
        \rowcolor{gray!20} 
        \multicolumn{16}{l}{\textit{Proprietary Model}}\\
        \textcolor{black!50}{GPT-4o} & & - & - & - & - & - & - & \textcolor{black!50}{18.8} & - & - & - & - & - & - & \textcolor{black!50}{20.1} \\
        \midrule
        \midrule
        \rowcolor{gray!20} 
        \multicolumn{16}{l}{\textit{Open-source Model}}\\
\textcolor{black!50}{Qwen2.5VL-3B~\cite{bai2025qwen2}} & & \textcolor{black!50}{90.2} & \textcolor{black!50}{72.9} & \textcolor{black!50}{78.4} & \textcolor{black!50}{57.1} & \textcolor{black!50}{80.9} & \textcolor{black!50}{64.1} & \textcolor{black!50}{72.1} & \textcolor{black!50}{88.5} & \textcolor{black!50}{74.4} & \textcolor{black!50}{78.9} & \textcolor{black!50}{58.6} & \textcolor{black!50}{76.9} & \textcolor{black!50}{64.5} & \textcolor{black!50}{73.6}\\
        \midrule
        \midrule
        \rowcolor{gray!20}
        \multicolumn{16}{l}{\textit{SFT-based GUI Method}}\\

        \multirow{1}{*}{\makecell[l]{}}
        & \textcolor{black!50}{Upper Bound}
        & \textcolor{black!50}{78.0}
        & \textcolor{black!50}{52.0}
        & \textcolor{black!50}{72.2}
        & \textcolor{black!50}{30.0}
        & \textcolor{black!50}{55.7}
        & \textcolor{black!50}{32.5}
        & \textcolor{black!50}{53.4}
        & \textcolor{black!50}{78.4}
        & \textcolor{black!50}{50.7}
        & \textcolor{black!50}{70.1}
        & \textcolor{black!50}{29.3}
        & \textcolor{black!50}{55.2}
        & \textcolor{black!50}{32.5}
        & \textcolor{black!50}{55.1} \\

        \multirow{3}{*}{SeeClick~\cite{cheng2024seeclick}}
        & M.
        & 52.7 & 59.8 & 52.1 & 42.9 & 20.4 & 25.7 & 42.3
        & 51.7 & 58.8 & 51.5 & 45.0 & 15.4 & 27.6 & 41.7 \\

        & M.$\rightarrow$D.
        & 69.1 & 63.4 & 79.8 & 48.5 & 67.2 & 44.3 & 62.1
        & 67.2 & 63.3 & 80.2 & 53.6 & 64.7 & 49.5 & 63.4 \\

        & M.$\rightarrow$D.$\rightarrow$W.
        & 73.9 & 65.5 & 78.4 & 57.9 & 69.1 & 53.4 & 66.4
        & 71.4 & 63.5 & 79.9 & 57.9 & 70.1 & 49.8 & 65.4 \\

        \midrule

        \multirow{1}{*}{\makecell[l]{}}
        & \textcolor{black!50}{Upper Bound}
        & \textcolor{black!50}{-}
        & \textcolor{black!50}{-}
        & \textcolor{black!50}{-}
        & \textcolor{black!50}{-}
        & \textcolor{black!50}{-}
        & \textcolor{black!50}{-}
        & \textcolor{black!50}{-}
        & \textcolor{black!50}{96.9}
        & \textcolor{black!50}{89.6}
        & \textcolor{black!50}{97.4}
        & \textcolor{black!50}{86.4}
        & \textcolor{black!50}{95.7}
        & \textcolor{black!50}{84.7}
        & \textcolor{black!50}{92.5} \\

        \multirow{3}{*}{GUI-Actor~\cite{wu2024atlas}}
        & M.
        & 74.4 & 76.4 & 58.8 & 61.4 & 32.2 & 63.1 & 61.0
        & 76.7 & 77.5 & 61.1 & 61.9 & 35.0 & 67.1 & 63.2 \\

        & M.$\rightarrow$D.
        & 91.2 & 71.2 & 85.1 & 70.0 & 74.4 & 69.4 & 76.9
        & 95.4 & 73.3 & 85.0 & 73.8 & 75.4 & 68.8 & 78.6 \\

        & M.$\rightarrow$D.$\rightarrow$W.
        & 87.6 & 70.7 & 87.1 & 67.1 & 72.6 & 65.1 & 75.0
        & 91.1 & 74.9 & 87.8 & 72.2 & 80.0 & 61.2 & 77.9 \\

        \midrule
        \midrule
        \rowcolor{gray!20}
        \multicolumn{16}{l}{\textit{RFT-based GUI Model}}\\

                    \multirow{1}{*}{\makecell[l]{\\}} 
                    & \textcolor{black!50}{Upper Bound} & \textcolor{black!50}{-} & \textcolor{black!50}{-} & \textcolor{black!50}{-} & \textcolor{black!50}{-} & \textcolor{black!50}{-} & \textcolor{black!50}{-} & \textcolor{black!50}{88.2} & \textcolor{black!50}{-} & \textcolor{black!50}{-} & \textcolor{black!50}{-} & \textcolor{black!50}{-} & \textcolor{black!50}{-} & \textcolor{black!50}{-} & \textcolor{black!50}{90.3} \\
        \multirow{3}{*}{\makecell[l]{SE-GUI$^{\dagger}$}~\cite{yuan2025enhancing}} & M. & 91.5 & 75.9 & 72.9 & 59.3 & 73.9 & 55.1 & 71.4 & 90.4 & 81.0 & 84.1 & 60.1 & 71.8 & 64.5 & 75.3\\ 
         & M.$\rightarrow$D. & 91.8 & 76.6 & 78.1 & 65.0 & 76.3 & 58.3 & 74.4 & 92.2 & 81.9 & 87.6 & 67.4 & 73.7 & 63.5 & 77.7\\
         & M.$\rightarrow$D.$\rightarrow$W. & 92.0 & 78.2 & 83.1 & 65.7 & 77.4 & 60.2 & 76.1 & 93.2 & 82.5 & 90.2 & 69.3 & 75.1 & 63.1 & 78.9\\
        \midrule 
                \multirow{1}{*}{\makecell[l]{}}
                    & \textcolor{black!50}{Upper Bound} & \textcolor{black!50}{96.7} & \textcolor{black!50}{90.8} & \textcolor{black!50}{95.9} & \textcolor{black!50}{88.6} & \textcolor{black!50}{90.9} & \textcolor{black!50}{86.9} & \textcolor{black!50}{92.0} & \textcolor{black!50}{-} & \textcolor{black!50}{-} & \textcolor{black!50}{-} & \textcolor{black!50}{-} & \textcolor{black!50}{-} & \textcolor{black!50}{-} & \textcolor{black!50}{93.3} \\
        \multirow{3}{*}{\makecell[l]{GUI-G$^{2}$~\cite{tang2025gui}}}
        & M. & 91.9 & 77.2 & 78.8 & 61.4 & 72.6 & 51.2 & 72.2 & 92.5 & 81.5 & 85.2 & 66.4 & 73.5 & 64.8 & 77.3 \\ 
         & M.$\rightarrow$D. & 93.8 & 76.4 & 88.2 & 61.4 & 75.2 & 53.9 & 74.8 & 94.2 & 78.7 & 87.2 & 66.4 & 77.8 & 65.7 & 78.3\\
         & M.$\rightarrow$D.$\rightarrow$W. & 95.2 & 76.4 & 92.8 & 63.6 & 79.6 & 54.9 & 77.1 & 94.8 & 79.1 & 89.3 & 68.6 & 81.2 & 68.6 & 80.3\\
        \midrule
        
        \multirow{3}{*}{GUI-AiF~\cite{guiaif}} & M. & 92.9 & 78.7 & 81.9 & 68.6 & 81.6 & 65.1 & 78.1 & 93.3 & 82.1 & 88.5 & 73.6 & 80.3 & 68.0 & 80.9 \\ 
         & M.$\rightarrow$D. & 94.1 & 77.3 & 93.2 & 69.3 & 81.9 & 66.2 & 80.3 & 95.9 & 79.6 & 89.7 & 77.1 & 81.2 & 68.3 & 81.9 \\
         & M.$\rightarrow$D.$\rightarrow$W. & 96.1 & 76.9 &  \textbf{95.7} & 68.3 & 84.8 & 68.2 & 81.7 & 96.6 &  \textbf{83.6} & 90.2 &  77.4 & 82.9 & 70.5 & 83.5 \\
        \midrule
        \multirow{3}{*}{\textbf{GUI-\textcolor[HTML]{C71030}{A}\textcolor[HTML]{3170B5}{C} (Ours)}} 
        & M. & 94.1 & 81.2 & 86.6 & 70.0 & 82.2 & 70.0 & 80.7 & 95.1 & 82.1 & 89.2 & 75.0 & 82.5 & 68.5 & 82.1 \\
        & M.$\rightarrow$D. & 95.2 & 81.2 & 92.2 & 70.7 & 87.8 & 69.0 & 82.7 & 96.5 & 82.5 & 92.3 & 77.1 & 83.7 & 68.5 & 83.4 \\
        & M.$\rightarrow$D.$\rightarrow$W. & \textbf{96.5} & \textbf{82.4} & 94.3 & \textbf{71.4} & \textbf{88.7} & \textbf{70.9} & \textbf{84.0} & \textbf{97.6}& 82.5 & \textbf{95.4} & \textbf{77.4} & \textbf{86.3} & \textbf{71.4} & \textbf{85.1} \\
        \midrule
        \bottomrule
    \end{tabular}
    }
\end{table*}

\begin{table*}[h]
\centering
\caption{Continual resolution performance (\%) of the proposed GUI-AC method on the ScreenSpot-Pro benchmark. The N. and H. denote the Normal and High resolution tasks, respectively. The \textcolor{gray}{upper bound} corresponds to the state-of-the-art performance achieved by RFT-based methods for GUI agents and is provided as a reference.}
\label{screenspotpro}
\resizebox{\textwidth}{!}{
\begin{tabular}{@{}ll* {12}{c}c@{}} 
  \toprule
\multirow{3}{*}{\textbf{Method}} & & \multicolumn{13}{c}{\textbf{SSPro Accuracy (\%)}} \\
    
    \cmidrule(lr){3-14} 
    
     & & \multicolumn{2}{c}{CAD} & \multicolumn{2}{c}{Dev} & \multicolumn{2}{c}{Creative} & \multicolumn{2}{c}{Scientific} & \multicolumn{2}{c}{Office} & \multicolumn{2}{c}{OS} & \textbf{Avg.} \\
    \cmidrule(lr){3-4} \cmidrule(lr){5-6} \cmidrule(lr){7-8} \cmidrule(lr){9-10} \cmidrule(lr){11-12} \cmidrule(lr){13-14}
    
     & & Text & Icon & Text & Icon & Text & Icon & Text & Icon & Text & Icon & Text & Icon & \\
 
  \midrule
  \rowcolor{gray!20}
  \multicolumn{15}{l}{\textit{Proprietary Model}}\\
 
\textcolor{black!50}{GPT-4o} & & \textcolor{black!50}{2.0} & \textcolor{black!50}{0.0} & \textcolor{black!50}{1.3} & \textcolor{black!50}{0.0} & \textcolor{black!50}{1.0} & \textcolor{black!50}{0.0} & \textcolor{black!50}{2.1} & \textcolor{black!50}{0.0} & \textcolor{black!50}{1.1} & \textcolor{black!50}{0.0} & \textcolor{black!50}{0.0} & \textcolor{black!50}{0.0} & \textcolor{black!50}{0.8} \\

\textcolor{black!50}{Claude Computer Use} & & \textcolor{black!50}{14.5} & \textcolor{black!50}{3.7} & \textcolor{black!50}{22.0} & \textcolor{black!50}{3.9} & \textcolor{black!50}{25.9} & \textcolor{black!50}{3.4} & \textcolor{black!50}{33.9} & \textcolor{black!50}{15.8} & \textcolor{black!50}{30.1} & \textcolor{black!50}{16.3} & \textcolor{black!50}{11.0} & \textcolor{black!50}{4.5} & \textcolor{black!50}{17.1} \\
 
  \midrule
  \midrule
  \rowcolor{gray!20}
  \multicolumn{15}{l}{\textit{Open-source Model}}\\
 
\textcolor{black!50}{Qwen2.5VL-3B~\cite{bai2025qwen2}} & & \textcolor{black!50}{9.1} & \textcolor{black!50}{7.3} & \textcolor{black!50}{22.1} & \textcolor{black!50}{1.4} & \textcolor{black!50}{26.8} & \textcolor{black!50}{2.1} & \textcolor{black!50}{38.2} & \textcolor{black!50}{7.3} & \textcolor{black!50}{33.9} & \textcolor{black!50}{15.1} & \textcolor{black!50}{10.3} & \textcolor{black!50}{1.1} & \textcolor{black!50}{16.1} \\
 
  \midrule
  \midrule
  \rowcolor{gray!20}
  \multicolumn{15}{l}{\textit{SFT-based GUI Methods}}\\

  \multirow{1}{*}{\makecell[l]{}}
  & \textcolor{black!50}{Upper Bound}
  & \textcolor{black!50}{2.5}
  & \textcolor{black!50}{0.0}
  & \textcolor{black!50}{0.6}
  & \textcolor{black!50}{0.0}
  & \textcolor{black!50}{1.0}
  & \textcolor{black!50}{0.0}
  & \textcolor{black!50}{3.5}
  & \textcolor{black!50}{0.0}
  & \textcolor{black!50}{1.1}
  & \textcolor{black!50}{0.0}
  & \textcolor{black!50}{2.8}
  & \textcolor{black!50}{0.0}
  & \textcolor{black!50}{1.1} \\

  \multirow{2}{*}{SeeClick~\cite{cheng2024seeclick}}
  & N.
  & 23.4 & 3.1
  & 16.2 & 0.7
  & 12.1 & 2.1
  & 18.1 & 11.8
  & 29.4 & 7.5
  & 5.6 & 3.4
  & 11.1 \\

  & N.$\rightarrow$H.
  & 26.4 & 6.3
  & 25.3 & 0.0
  & 20.8 & 5.6
  & 28.1 & 14.6
  & 35.6 & 11.3
  & 17.7 & 8.9
  & 16.7 \\

  \midrule
  \midrule
  \rowcolor{gray!20}
  \multicolumn{15}{l}{\textit{RFT-based GUI Methods}}\\
  & \textcolor{black!50}{Upper Bound} & \textcolor{black!50}{55.8} & \textcolor{black!50}{12.5} &\textcolor{black!50}{68.8} & \textcolor{black!50}{17.2} & \textcolor{black!50}{57.1} & \textcolor{black!50}{15.4} &\textcolor{black!50}{77.1} & \textcolor{black!50}{24.5} & \textcolor{black!50}{74.0} & \textcolor{black!50}{32.7} &\textcolor{black!50}{57.9} & \textcolor{black!50}{21.3} & \textcolor{black!50}{47.5} \\
  \multirow{2}{*}{\makecell[l]{GUI-G$^{2}$~\cite{tang2025gui}}} 
  & N. & 24.3 & 1.7 & 17.5 & 2.1 & 20.2 & 6.3 & 22.2 & 12.8 & 36.2 & 11.3 & 17.7 & 4.5 & 14.7 \\
  & N.$\rightarrow$H. & 24.5 & 6.3 & 24.4 & 4.1 & 23.7 & \textbf{5.6} & 27.3 & 12.6 & 35.1 & 9.4 & 21.2 & 5.6 & 16.7 \\
 
  \midrule
  \multirow{2}{*}{GUI-AiF~\cite{guiaif}} & N. & 27.4 & 4.7 & 24.7 & 1.4 & 17.7 & 8.2 & 29.2 & 14.6 & 35.3 & 20.8 & 17.8 &  10.1 & 17.7 \\
  & N.$\rightarrow$H. & 20.3 & \textbf{15.6} & 29.2 & 2.1 & 23.2 & 3.5 & 33.3 & 14.7 & 38.4 & \textbf{18.2} & 22.4 & 6.8 & 19.0 \\
 
  \midrule
  \multirow{2}{*}{\textbf{GUI-\textcolor[HTML]{C71030}{A}\textcolor[HTML]{3170B5}{C} (Ours)}} 
  & N. 
  & 25.4 & 3.1 & 29.2 & 2.1 & 32.3 & 2.1 & 36.8 & 12.6 & 49.2 & 13.2 & 22.4 & 4.5 & 19.4 \\
  & N.$\rightarrow$H. 
  & \textbf{27.9} & 7.3 & \textbf{39.6} & \textbf{4.1} & \textbf{36.4} & \textbf{5.6} & \textbf{41.0} & \textbf{15.5} & \textbf{49.7} & 17.0 & \textbf{26.2} & \textbf{7.9} & \textbf{23.1} \\
  \midrule
  \bottomrule
\end{tabular}
}
\end{table*}

\section{Experiments} 
\label{sec:experiments}
\subsection{Experimental Setup}

\textbf{Dataset and Evaluation Benchmarks.}
We follow the continual learning evaluation protocol of Continual GUI Agents and evaluate under two sequential training regimes: (i) Cross-Domain transfer (Mobile $\rightarrow$ Desktop $\rightarrow$ Web) and (ii) Cross-Resolution transfer (Normal Web $\rightarrow$ High Web). The cross-domain stream captures heterogeneous interface distributions relevant to Web-oriented GUI agents, while the cross-resolution stream evaluates continual adaptation to resolution shifts in Web interfaces. We use Widget Captioning~\cite{cheng2024seeclick} for mobile applications, ShowUI-web~\cite{lin2025showui} for standard-resolution web applications (mostly 1080p), and OmniACT~\cite{kapoor2024omniact} for desktop and high-resolution Web interfaces (up to 4K). For cross-domain continual learning, we train sequentially on Widget Captioning (Mobile) and then on OmniACT (Desktop and Web). For cross-resolution continual learning, we train from ShowUI-web (Normal) to OmniACT-web (High). We evaluate our method on ScreenSpot-V1 (SSv1)~\cite{cheng2024seeclick}, ScreenSpot-V2 (SSv2)~\cite{wu2024atlas}, and ScreenSpot-Pro (SSPro)~\cite{li2025screenspot}. SSv1 and SSv2 cover mobile, desktop, and Web interfaces, enabling evaluation of cross-domain continual learning. SSPro focuses on high-resolution software interfaces and contains six categories: CAD, development, creative software, scientific and analytic software, office software, and operating systems. We use SSPro to evaluate cross-resolution transfer.

\textbf{Baseline Methods and Implementation Details.}
We primarily focus on comparisons with RFT-based methods, while including representative SFT-based methods as reference points. Specifically, we consider SeeClick~\cite{cheng2024seeclick} and GUI-Actor~\cite{wu2024atlas} as SFT-based references. For RFT-based methods, we compare against: InfiGUI-R1~\cite{liu2025infigui}, SE-GUI~\cite{yuan2025enhancing}, GUI-G$^2$~\cite{tang2025gui}, and GUI-AiF~\cite{guiaif}. InfiGUI-R1 uses the IoU between predicted and ground-truth bounding boxes as the primary feedback signal. SE-GUI uses the distance between box centers as the reward. GUI-G$^2$ models bounding boxes with a Gaussian distribution to construct a dense reward that jointly accounts for center deviation and region coverage. GUI-AiF improves robustness to domain and resolution shifts by incorporating point-level and region-level rewards. SE-GUI$^{\dagger}$ denotes a hybrid variant that combines the IoU reward from InfiGUI-R1 with the distance reward from SE-GUI to improve training stability.

We adopt Qwen2.5VL-3B~\cite{bai2025qwen2} as the vision--language backbone and use the same task format as GUI-AiF~\cite{guiaif} for a
controlled comparison. All models are trained with bfloat16; FlashAttention-2 and gradient checkpointing are enabled for efficiency. Unless otherwise stated, we use a learning rate of $1\times10^{-6}$, a global batch size of 8, and sample 4 candidate predictions per instruction for optimization. We set the KL regularization coefficient to $\beta=0.04$ and the base clipping margin to $\epsilon_0=0.2$. For GUI-AC, we additionally set $a_{\min}=0.2$ and $k=3.0$.

\begin{table*}[h] 
    \centering
    \caption{Ablation study of GUI-AC on ScreenSpot-V1 and ScreenSpot-V2 benchmark. The M. D. and W. denote the Mobile, Desktop and Web domain tasks, respectively.}
    \label{tab:singlereward}
    
    \resizebox{\textwidth}{!}{ 
    \begin{tabular}{@{}l l cc cc cc c cc cc cc c@{}} 
        \toprule
        
        \multirow{3.5}{*}{\textbf{Method}} & \multirow{3.5}{*} & \multicolumn{7}{c}{\textbf{SSv1 Accuracy (\%)}} & \multicolumn{7}{c}{\textbf{SSv2 Accuracy (\%)}} \\
        
        \cmidrule(lr){3-9} \cmidrule(lr){10-16}
         & & \multicolumn{2}{c}{Mobile} & \multicolumn{2}{c}{Desktop} & \multicolumn{2}{c}{Web} & \multirow{2.5}{*}{\textbf{Avg.}} & \multicolumn{2}{c}{Mobile} & \multicolumn{2}{c}{Desktop} & \multicolumn{2}{c}{Web} & \multirow{2.5}{*}{\textbf{Avg.}} \\
        
        \cmidrule(lr){3-4} \cmidrule(lr){5-6} \cmidrule(lr){7-8} \cmidrule(lr){10-11} \cmidrule(lr){12-13} \cmidrule(lr){14-15}
         & & Text & Icon & Text & Icon & Text & Icon & & Text & Icon & Text & Icon & Text & Icon & \\
        
        \midrule
        
        \multirow{3}{*}{\textbf{GUI-\textcolor[HTML]{C71030}{A}\textcolor[HTML]{3170B5}{C}}} 
        & M. & 94.1 & 81.2 & 86.6 & 70.0 & 82.2 & 70.0 & 80.7 & 95.1 & 82.1 & 89.2 & 75.0 & 82.5 & 68.5 & 82.1 \\
        & M.$\rightarrow$D. & 95.2 & 81.2 & 92.2 & 70.7 & 87.8 & 69.0 & 82.7 & 96.5 & 82.5 & 92.3 & 77.1 & 83.7 & 68.5 & 83.4 \\
        & M.$\rightarrow$D.$\rightarrow$W. & \textbf{96.5} & \textbf{82.4} & \textbf{94.3} & \textbf{71.4} & \textbf{88.7} & \textbf{70.9} & \textbf{84.0} & \textbf{97.6} & \textbf{82.5} & \textbf{95.4} & \textbf{77.4} & \textbf{86.3} & \textbf{71.4} & \textbf{85.1} \\
        
        \midrule
        \midrule
        
        \multirow{3}{*}{\makecell[l]{GUI-\textbf{\textcolor[HTML]{C71030}{A}}}} & M. &
        93.8 & 81.7 & 86.1 & 64.3 & 83.3 & 68.2 & 79.6 &
        94.8 & 83.9 & 87.1 & 68.6 & 84.1 & 70.9 & 81.6 \\
        
         & M.$\rightarrow$D. &
        94.1 & 80.6 & 91.2 & 68.6 & 84.2 & 70.0 & 81.5 &
        95.9 & 81.7 & 91.8 & 73.6 & 81.2 & 68.6 & 82.1 \\
        
         & M.$\rightarrow$D.$\rightarrow$W. &
        96.2 & 78.6 & 93.7 & 68.6 & 84.9 & 70.7 & 82.1 &
        97.2 & 82.1 & 92.8 & 75.0 & 83.7 & 70.5 & 83.6 \\
        
        \midrule
        \midrule
        
        \multirow{3}{*}{\makecell[l]{GUI-\textbf{\textcolor[HTML]{3170B5}{C}}}} & M. &
        94.1 & 80.6 & 88.2 & 63.6 & 82.2 & 68.6 & 79.6
        & 95.5 & 82.1 & 88.4 & 70.0 & 83.8 & 70.1 & 81.7 \\
        
         & M.$\rightarrow$D. &
        95.6 & 81.2 & 91.2 & 66.4 & 84.8 & 69.0 & 81.4  &
        96.2 & 81.5 & 92.3 & 73.6 & 81.2 & 68.3 & 82.2 \\
        
         & M.$\rightarrow$D.$\rightarrow$W. &
        96.3 & 81.6 & 92.3 & 68.6 & 86.2 & 69.0 & 82.3 &
        97.2 & 82.0 & 93.2 & 77.1 & 84.3 & 70.8 & 84.1 \\
        
        \midrule
        \bottomrule
    \end{tabular}
    } 
\end{table*}

\begin{figure*}[h]
  \centering
  \includegraphics[width=0.9\linewidth]{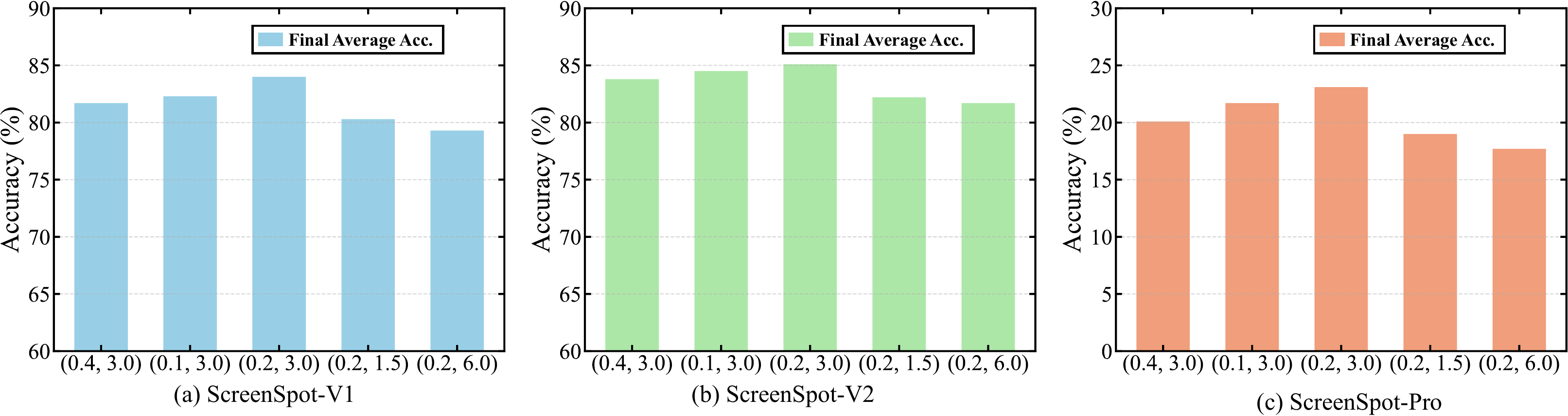}
  \caption{Hyperparameter sensitivity analysis for $a_{\min}$ and $k$. Performance peaks at $(a_{\min}=0.2,\,k=3.0)$ on ScreenSpot-V1, ScreenSpot-V2 and ScreenSpot-Pro benchmarks.}
  \label{fig:hyperparam}
\end{figure*}

\subsection{Main Results}

\textbf{Continual GUI Domain.}
We first evaluate the continual-domain performance of GUI-AC. Table~\ref{screenspot12} shows that GUI-AC delivers consistent improvements under continual domain shifts. In line with prior observations in continual learning of GUI agents, proprietary general-purpose models exhibit low grounding accuracy. Open-source VLM backbones provide a stronger starting point, but still leave substantial headroom. Among adaptation strategies, GUI-AC achieves the highest accuracy, with stable gains for both text-based and icon-based interactions across both SSv1 and SSv2. Importantly, performance accumulates over time: as training proceeds along the cross-domain stream, average accuracy increases steadily, suggesting that learning on later domains does not substantially erode earlier grounding skills and instead yields net improvements across domains. We further validate the robustness of GUI-AC across three independent random seeds (Appendix~\ref{app:multiseed}). GUI-AC also achieves stronger forward and backward transfer than
GUI-AiF (Appendix~\ref{app:continual_metrics}). Additional comparisons with stronger RL optimization baselines (Appendix~\ref{app:rlbaseline}) and experiments under a reversed training sequence (Appendix~\ref{app:reverse_order}) show that the improvements remain consistent across alternative optimization strategies and different arrival orders.


\textbf{Continual GUI Resolution.}
We then evaluate the continual-resolution performance of GUI-AC. Table~\ref{screenspotpro} further indicates that GUI-AC is robust to continual resolution shifts and exhibits clear forward transfer. Since the Gaussian modeling-based RFT provides better performance, we compare against GUI-G$^{2}$ and GUI-AiF as representative RFT baselines throughout these evaluations. Overall, GUI-AC performs best under continual resolution shifts. After continual training on high-resolution data, accuracy improves across all categories, with the largest gains on text targets. Across methods, icon grounding remains more difficult than text grounding, likely because small icons provide limited visual evidence in dense professional interfaces. Nevertheless, GUI-AC consistently improves icon performance as well, demonstrating stronger robustness under continual resolution variation.

\subsection{Ablation Study}

\textbf{Certainty-Calibrated Components.} 
We conduct an ablation study of GUI-AC. As shown in Table~\ref{tab:singlereward}, we isolate the two certainty-calibrated components by activating either Adaptive Advantage (GUI-A) or Dynamic Clipping (GUI-C) alone. Overall, the full GUI-AC yields the highest average accuracy, while removing either component consistently degrades performance, indicating that the two mechanisms provide complementary benefits. Moreover, GUI-C achieves a higher final average accuracy than GUI-A on both SSV1 and SSV2, implying that Dynamic Clipping is the primary driver of the improvement. These results suggest that fixed clipping imposes a stronger constraint on exploration, and that alleviating this constraint is crucial for performance.

\textbf{Hyperparameter Sensitivity.}
We further analyze the sensitivity of GUI-AC to two key hyperparameters: $a_{\min}$ and $k$. Figure~\ref{fig:hyperparam} summarizes the final average accuracy across different $(a_{\min}, k)$ configurations. Notably, the majority of parameter configurations still yield better performance than GUI-AIF. Among the values tested, the default choice $(0.2, 3.0)$ performs best on all benchmarks. A larger $a_{\min}$ strengthens the negative correction for overconfident rollouts, whereas an excessively large value may over-amplify noisy negative updates. Conversely, a small $a_{\min}$ provides insufficient correction for overconfidence. The expansion factor $k$ exhibits a similar trade-off. If $k$ is too small, the trust region is not sufficiently enlarged under uncertainty and if $k$ is too large, the resulting permissive exploration can exacerbate noisy credit assignment. Taken together, GUI-AC remains stable under moderate hyperparameter variations but benefits from a balanced choice of $(a_{\min}, k)$.

\textbf{Grounding Certainty Sensitivity.}
We further evaluate the sensitivity to the threshold $\tau$ by testing additional values (0.4 and 0.8) alongside the default 0.6. The results, as shown in Table~\ref{tab:groundingcertainty}, both 0.4 and 0.8 underperform the default value of 0.6, indicating that a moderate threshold achieves a better balance between filtering noisy signals and preserving useful training samples.

\begin{table*}[h] 
    \centering
    \caption{Grounding certainty sensitivity study to the threshold on ScreenSpot-V1 and ScreenSpot-V2 benchmark. The M. D. and W. denote the Mobile, Desktop and Web domain tasks, respectively.}
    \label{tab:groundingcertainty}
    
    \resizebox{\textwidth}{!}{ 
    \begin{tabular}{@{}l l cc cc cc c cc cc cc c@{}} 
        \toprule
        
        \multirow{3.5}{*}{\textbf{Method}} & \multirow{3.5}{*} & \multicolumn{7}{c}{\textbf{SSv1 Accuracy (\%)}} & \multicolumn{7}{c}{\textbf{SSv2 Accuracy (\%)}} \\
        
        \cmidrule(lr){3-9} \cmidrule(lr){10-16}
         & & \multicolumn{2}{c}{Mobile} & \multicolumn{2}{c}{Desktop} & \multicolumn{2}{c}{Web} & \multirow{2.5}{*}{\textbf{Avg.}} & \multicolumn{2}{c}{Mobile} & \multicolumn{2}{c}{Desktop} & \multicolumn{2}{c}{Web} & \multirow{2.5}{*}{\textbf{Avg.}} \\
        
        \cmidrule(lr){3-4} \cmidrule(lr){5-6} \cmidrule(lr){7-8} \cmidrule(lr){10-11} \cmidrule(lr){12-13} \cmidrule(lr){14-15}
         & & Text & Icon & Text & Icon & Text & Icon & & Text & Icon & Text & Icon & Text & Icon & \\
        
        \midrule
        
        \multirow{3}{*}{\textbf{$\tau=0.6$}} 
        & M. & 94.1 & 81.2 & 86.6 & 70.0 & 82.2 & 70.0 & 80.7 & 95.1 & 82.1 & 89.2 & 75.0 & 82.5 & 68.5 & 82.1 \\
        & M.$\rightarrow$D. & 95.2 & 81.2 & 92.2 & 70.7 & 87.8 & 69.0 & 82.7 & 96.5 & 82.5 & 92.3 & 77.1 & 83.7 & 68.5 & 83.4 \\
        & M.$\rightarrow$D.$\rightarrow$W. & \textbf{96.5} & \textbf{82.4} & \textbf{94.3} & \textbf{71.4} & \textbf{88.7} & \textbf{70.9} & \textbf{84.0} & \textbf{97.6} & \textbf{82.5} & \textbf{95.4} & \textbf{77.4} & \textbf{86.3} & \textbf{71.4} & \textbf{85.1} \\
        
        \midrule
        \midrule
        
        \multirow{3}{*}{\makecell[l]{\textbf{$\tau=0.4$}}} & M. &
        94.0 & 81.0 & 86.1 & 69.3 & 81.8 & 69.4 & 80.3 & 94.9 & 81.8 & 88.8 & 74.3 & 82.0 & 68.0 & 81.6 \\
        
         & M.$\rightarrow$D. &
        95.0 & 80.9 & 91.6 & 70.0 & 87.0 & 68.2 & 82.1 & 96.2 & 82.1 & 91.8 & 76.4 & 83.0 & 68.0 & 82.9 \\
        
         & M.$\rightarrow$D.$\rightarrow$W. &
        96.2 & 81.9 & 93.8 & 70.8 & 88.0 & 70.2 & 83.5 & 97.3 & 82.1 & 94.8 & 76.6 & 85.4 & 70.8 & 84.5 \\
        
        \midrule
        \midrule
        
        \multirow{3}{*}{\makecell[l]{\textbf{$\tau=0.8$}}} & M. &
        93.6 & 80.4 & 85.3 & 68.7 & 80.9 & 68.5 & 79.6 & 
        94.4 & 81.3 & 87.9 & 73.5 & 81.1 & 67.0 & 80.9 \\
        
         & M.$\rightarrow$D. &
        94.7 & 80.2 & 90.9 & 69.2 & 85.9 & 67.4 & 81.4 & 
        95.8 & 81.6 & 90.8 & 75.6 & 82.1 & 67.2 & 82.2 \\
        
         & M.$\rightarrow$D.$\rightarrow$W. &
        95.8 & 81.0 & 92.9 & 70.0 & 86.7 & 69.1 & 82.6 & 
        96.8 & 81.8 & 93.6 & 76.0 & 84.6 & 69.8 & 83.8 \\
        
        \midrule
        \bottomrule
    \end{tabular}
    } 
\end{table*}

\subsection{Discussion}

\begin{figure}[h]
  \centering
  \includegraphics[width=\linewidth]{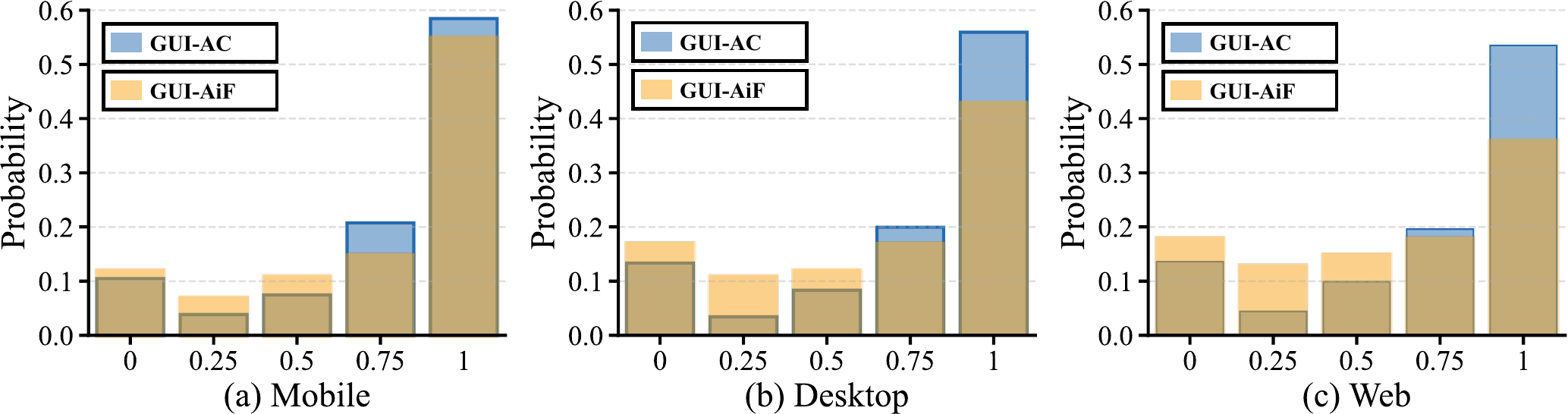}
  \caption{Histograms of grounding certainty under $n=4$ sampled candidates per instruction.} 
  \label{fig:certainty}
\end{figure}

\textbf{Grounding Certainty Distribution Analysis.}
Figure~\ref{fig:certainty} characterizes how grounding certainty evolves under cross-domain continual learning by comparing GUI-AiF and GUI-AC. Because we sample $n=4$ candidates per instruction, the certainty score is quantized to $\{0, 0.25, 0.5, 0.75, 1\}$. In the Mobile stage, the distribution places most of its mass in high-certainty bins, suggesting stable grounding when the domain is unchanged. As the interface shifts from Mobile to Desktop and further to Web, the distribution progressively rebalances toward lower-certainty bins, indicating growing uncertainty induced by domain drift. This effect is substantially stronger for GUI-AiF. In contrast, GUI-AC exhibits a gentler redistribution and retains more high-certainty mass, implying improved robustness of grounding under continual domain shifts.

\begin{figure}[t]
    \centering
    \includegraphics[width=\linewidth]{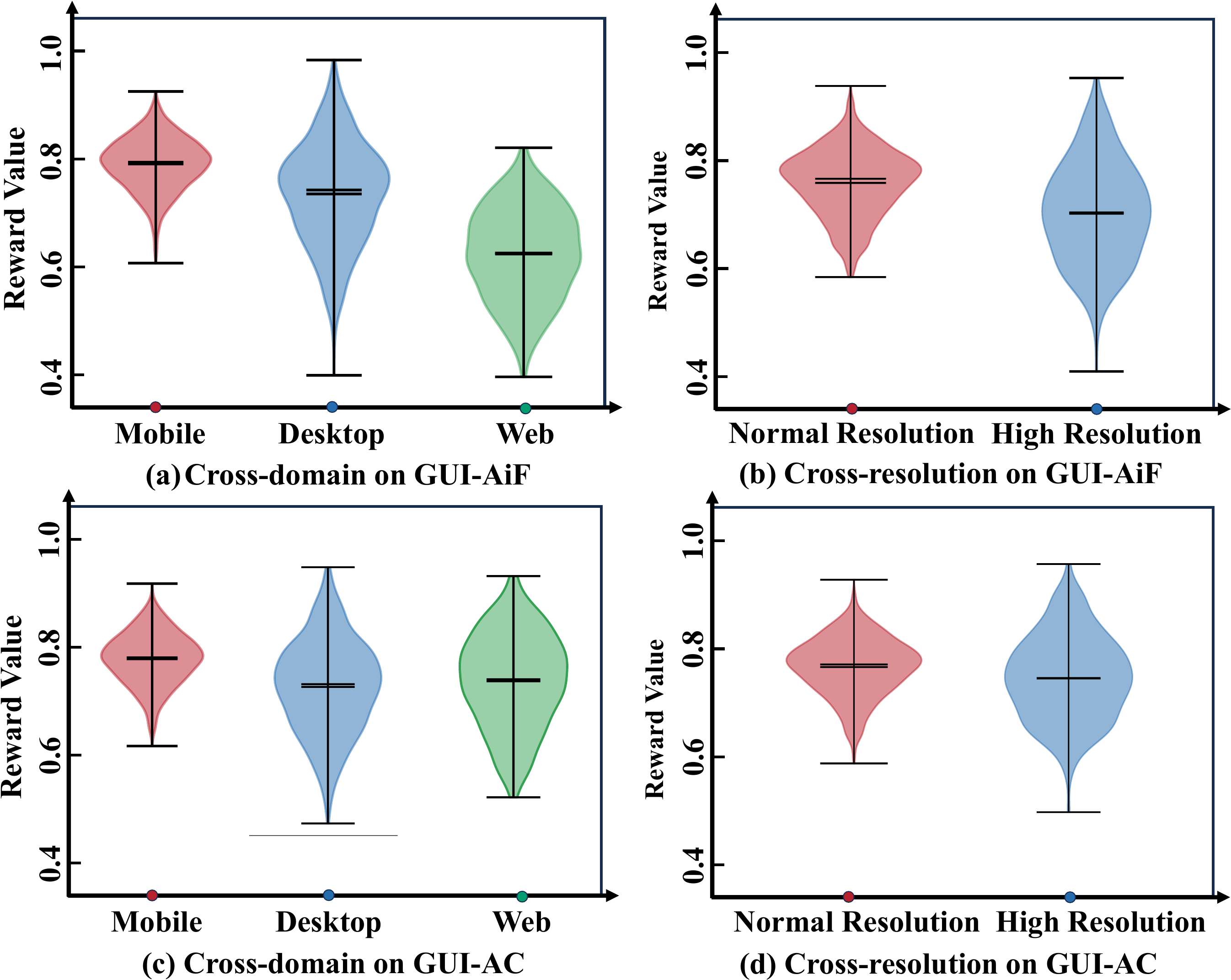}
    \caption{Reward distributions under continual learning of GUI agents. (a) and (c) show cross-domain settings, while (b) and (d) show cross-resolution settings.}
    \label{fig:distribution}
    \vspace{-3mm}
\end{figure}

\textbf{Reward Distribution Analysis.}
Figure~\ref{fig:distribution} presents violin plots of stage-wise reward distributions to assess optimization stability under continual GUI shifts. Each violin summarizes the distribution of per-rollout rewards within a stage. We observe consistent patterns in both the cross-domain and cross-resolution evaluations. Under GUI-AiF, rewards become increasingly dispersed and develop a pronounced lower tail after each transfer, suggesting deteriorating stability. By contrast, GUI-AC yields markedly more stable distributions across the stream: it suppresses tail growth, limits distributional widening after shifts, and preserves the bulk of the reward mass. Overall, these results support that GUI-AC enhances training stability for continual learning of GUI agents.

\begin{figure}[t]
    \centering
    \includegraphics[width=\linewidth]{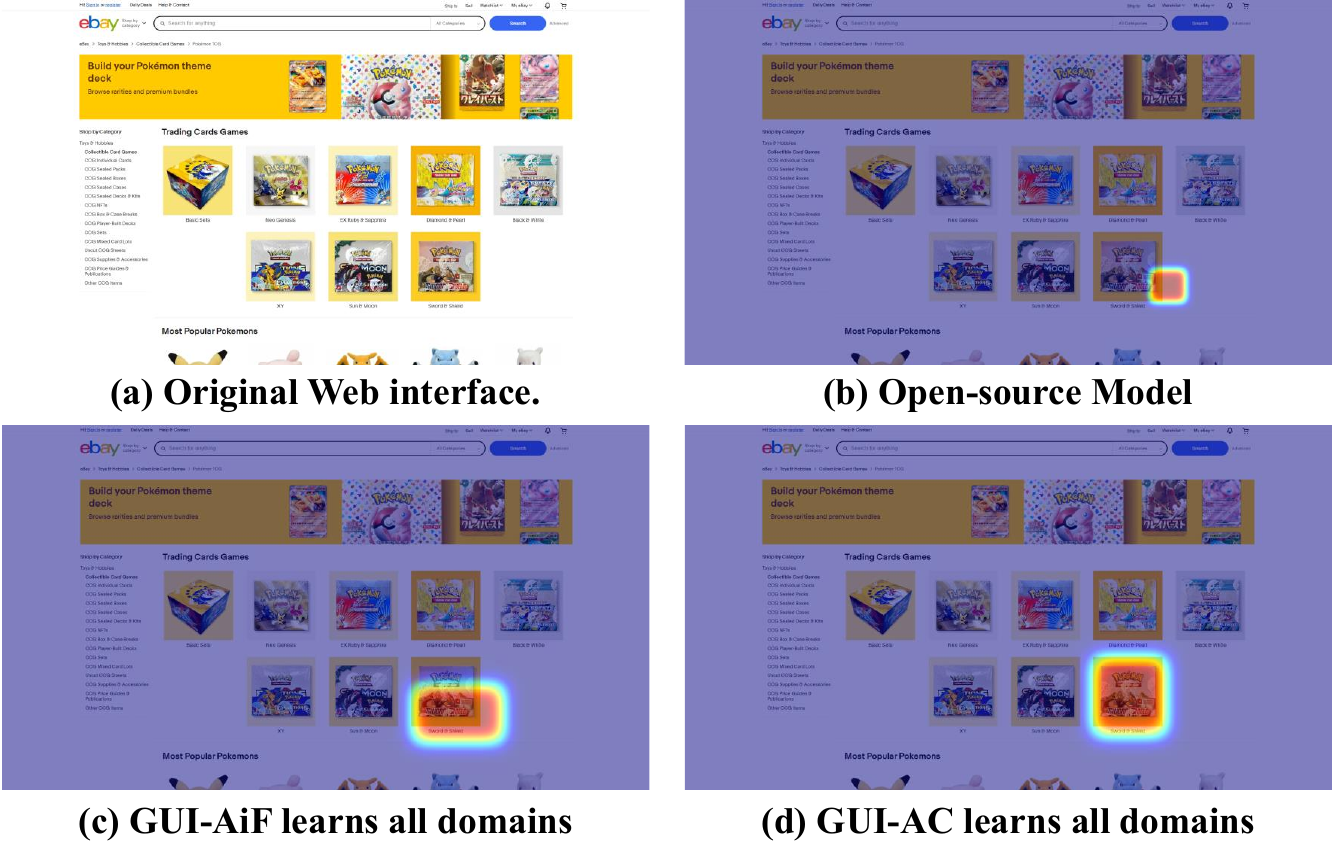}
    \caption{Visualization of interaction-region heatmaps on Web platforms. The instruction is: \emph{``View more details about the sword and shield item.''}}
    \label{fig:Visualization}
    \vspace{-6mm}
\end{figure}

\textbf{Visualization Comparison.}
Figure~\ref{fig:Visualization} compares interaction-region heatmaps in terms of spatial alignment and concentration. Figure~\ref{fig:Visualization}a displays the Web interface with the instruction ``View more details about the sword and shield item''. The open-source baseline shows weak, poorly aligned activation, where the peak often shifts to nearby background regions (Fig.~\ref{fig:Visualization}b). GUI-AiF amplifies target-related activation, yet retains noticeable spillover to adjacent candidates (Fig.~\ref{fig:Visualization}c). By contrast, GUI-AC produces a sharply localized peak on the correct item with minimal surrounding mass (Fig.~\ref{fig:Visualization}d), indicating more accurate and robust GUI grounding.

\section{Conclusion}

In this paper, we further investigated the continual learning of GUI Agents. We reveal the limitations of unreliable advantage estimation and fixed clipping under non-stationary GUI data. Then, we proposed GUI-AC, which leverages adaptive advantage and dynamic clipping to enhance continual learning performance of GUI agents. Extensive experiments demonstrate that GUI-AC consistently adapts to varying GUI domain and resolutions, outperforming baselines in dynamic GUI environments.

\begin{acks}
This work was supported by Alibaba Group through the Alibaba Innovative Research Program, the National Key Research and Development Program of China (Grant No. 2024YFC3308500), the Beijing Municipal Natural Science Foundation (Grant No. L251042), and the National Natural Science Foundation of China (Grant No. 62406036). This work was also supported by the State Key Laboratory of Networking and Switching Technology (Grant No. NST20250110).
\end{acks}

\bibliographystyle{ACM-Reference-Format}
\bibliography{Reference}

@String{Computing = "Computing" }

@String{Computer = "{IEEE} Computer" }

@String{Springer = "Springer-Verlag" }

@article{wang2024gui,
  title={{GUI Agents with Foundation Models: A Comprehensive Survey}},
  author={Wang, Shuai and Liu, Weiwen and Chen, Jingxuan and Zhou, Yuqi and Gan, Weinan and Zeng, Xingshan and Che, Yuhan and Yu, Shuai and Hao, Xinlong and Shao, Kun and others},
  journal={arXiv preprint arXiv:2411.04890},
  year={2024}
}

@article{tang2025survey,
  title={{A Survey on (M)LLM-Based GUI Agents}},
  author={Tang, Fei and Xu, Haolei and Zhang, Hang and Chen, Siqi and Wu, Xingyu and Shen, Yongliang and Zhang, Wenqi and Hou, Guiyang and Tan, Zeqi and Yan, Yuchen and others},
  journal={arXiv preprint arXiv:2504.13865},
  year={2025}
}

@inproceedings{hong2024cogagent,
  title={{CogAgent: A Visual Language Model for GUI Agents}},
  author={Hong, Wenyi and Wang, Weihan and Lv, Qingsong and Xu, Jiazheng and Yu, Wenmeng and Ji, Junhui and Wang, Yan and Wang, Zihan and Dong, Yuxiao and Ding, Ming and others},
  booktitle={Proceedings of the IEEE/CVF Conference on Computer Vision and Pattern Recognition},
  pages={14281--14290},
  year={2024}
}

@inproceedings{guan2025kg,
  title={{KG-RAG: Enhancing GUI Agent Decision-Making via Knowledge Graph-Driven Retrieval-Augmented Generation}},
  author={Guan, Ziyi and Li, Jason Chun Lok and Hou, Zhijian and Zhang, Pingping and Xu, Donglai and Zhao, Yuzhi and Wu, Mengyang and Chen, Jinpeng and Nguyen, Thanh-Toan and Xian, Pengfei and others},
  booktitle={Proceedings of the 2025 Conference on Empirical Methods in Natural Language Processing},
  pages={5396--5405},
  year={2025}
}

@inproceedings{cheng2024seeclick,
  title={{SeeClick: Harnessing GUI Grounding for Advanced Visual GUI Agents}},
  author={Cheng, Kanzhi and Sun, Qiushi and Chu, Yougang and Xu, Fangzhi and YanTao, Li and Zhang, Jianbing and Wu, Zhiyong},
  booktitle={Proceedings of the 62nd Annual Meeting of the Association for Computational Linguistics (Volume 1: Long Papers)},
  pages={9313--9332},
  year={2024}
}

@inproceedings{lin2025showui,
  title={{ShowUI: One Vision-Language-Action Model for GUI Visual Agent}},
  author={Lin, Kevin Qinghong and Li, Linjie and Gao, Difei and Yang, Zhengyuan and Wu, Shiwei and Bai, Zechen and Lei, Stan Weixian and Wang, Lijuan and Shou, Mike Zheng},
  booktitle={Proceedings of the Computer Vision and Pattern Recognition Conference},
  pages={19498--19508},
  year={2025}
}

@inproceedings{feng2022overcoming,
  title={{Overcoming Catastrophic Forgetting in Incremental Object Detection via Elastic Response Distillation}},
  author={Feng, Tao and Wang, Mang and Yuan, Hangjie},
  booktitle={Proceedings of the IEEE/CVF conference on computer vision and pattern recognition},
  pages={9427--9436},
  year={2022}
}

@article{du2025test,
  title={{Test-Time Reinforcement Learning for GUI Grounding via Region Consistency}},
  author={Du, Yong and Yan, Yuchen and Tang, Fei and Lu, Zhengxi and Zong, Chang and Lu, Weiming and Jiang, Shengpei and Shen, Yongliang},
  journal={arXiv preprint arXiv:2508.05615},
  year={2025}
}

@article{gou2024navigating,
  title={{Navigating the Digital World as Humans Do: Universal Visual Grounding for GUI Agents}},
  author={Gou, Boyu and Wang, Ruohan and Zheng, Boyuan and Xie, Yanan and Chang, Cheng and Shu, Yiheng and Sun, Huan and Su, Yu},
  journal={arXiv preprint arXiv:2410.05243},
  year={2024}
}

@article{zhang2024large,
  title={{Large Language Model-Brained GUI Agents: A Survey}},
  author={Zhang, Chaoyun and He, Shilin and Qian, Jiaxu and Li, Bowen and Li, Liqun and Qin, Si and Kang, Yu and Ma, Minghua and Liu, Guyue and Lin, Qingwei and others},
  journal={arXiv preprint arXiv:2411.18279},
  year={2024}
}

@inproceedings{sun2025genesis,
  title={{OS-Genesis: Automating GUI Agent Trajectory Construction via Reverse Task Synthesis}},
  author={Sun, Qiushi and Cheng, Kanzhi and Ding, Zichen and Jin, Chuanyang and Wang, Yian and Xu, Fangzhi and Wu, Zhenyu and Jia, Chengyou and Chen, Liheng and Liu, Zhoumianze and others},
  booktitle={Proceedings of the 63rd Annual Meeting of the Association for Computational Linguistics (Volume 1: Long Papers)},
  pages={5555--5579},
  year={2025}
}

@inproceedings{yang2025aria,
  title={{Aria-UI: Visual Grounding for GUI Instructions}},
  author={Yang, Yuhao and Wang, Yue and Li, Dongxu and Luo, Ziyang and Chen, Bei and Huang, Chao and Li, Junnan},
  booktitle={Findings of the Association for Computational Linguistics: ACL 2025},
  pages={22418--22433},
  year={2025}
}

@article{gao2025uishift,
  title={{UIShift: Enhancing VLM-based GUI Agents through Self-supervised Reinforcement Learning}},
  author={Gao, Longxi and Zhang, Li and Xu, Mengwei},
  journal={arXiv preprint arXiv:2505.12493},
  year={2025}
}

@article{tang2025gui,
  title={{GUI-G$^2$: Gaussian Reward Modeling for GUI Grounding}},
  author={Tang, Fei and Gu, Zhangxuan and Lu, Zhengxi and Liu, Xuyang and Shen, Shuheng and Meng, Changhua and Wang, Wen and Zhang, Wenqi and Shen, Yongliang and Lu, Weiming and others},
  journal={arXiv preprint arXiv:2507.15846},
  year={2025}
}

@article{wang2024mobile,
  title={{Mobile-Agent: Autonomous Multi-Modal Mobile Device Agent with Visual Perception}},
  author={Wang, Junyang and Xu, Haiyang and Ye, Jiabo and Yan, Ming and Shen, Weizhou and Zhang, Ji and Huang, Fei and Sang, Jitao},
  journal={arXiv preprint arXiv:2401.16158},
  year={2024}
}

@inproceedings{zhang2025ufo,
  title={{UFO: A UI-Focused Agent for Windows OS Interaction}},
  author={Zhang, Chaoyun and Li, Liqun and He, Shilin and Zhang, Xu and Qiao, Bo and Qin, Si and Ma, Minghua and Kang, Yu and Lin, Qingwei and Rajmohan, Saravan and others},
  booktitle={Proceedings of the 2025 Conference of the Nations of the Americas Chapter of the Association for Computational Linguistics: Human Language Technologies (Volume 1: Long Papers)},
  pages={597--622},
  year={2025}
}

@article{liu2024autoglm,
  title={{AutoGLM: Autonomous Foundation Agents for GUIs}},
  author={Liu, Xiao and Qin, Bo and Liang, Dongzhu and Dong, Guang and Lai, Hanyu and Zhang, Hanchen and Zhao, Hanlin and Iong, Iat Long and Sun, Jiadai and Wang, Jiaqi and others},
  journal={arXiv preprint arXiv:2411.00820},
  year={2024}
}

@article{wu2024atlas,
  title={{OS-ATLAS: A Foundation Action Model for Generalist GUI Agents}},
  author={Wu, Zhiyong and Wu, Zhenyu and Xu, Fangzhi and Wang, Yian and Sun, Qiushi and Jia, Chengyou and Cheng, Kanzhi and Ding, Zichen and Chen, Liheng and Liang, Paul Pu and others},
  journal={arXiv preprint arXiv:2410.23218},
  year={2024}
}

@inproceedings{zhang2025appagent,
  title={{AppAgent: Multimodal Agents as Smartphone Users}},
  author={Zhang, Chi and Yang, Zhao and Liu, Jiaxuan and Li, Yanda and Han, Yucheng and Chen, Xin and Huang, Zebiao and Fu, Bin and Yu, Gang},
  booktitle={Proceedings of the 2025 CHI Conference on Human Factors in Computing Systems},
  pages={1--20},
  year={2025}
}

@inproceedings{kirillov2023segment,
  title={{Segment Anything}},
  author={Kirillov, Alexander and Mintun, Eric and Ravi, Nikhila and Mao, Hanzi and Rolland, Chloe and Gustafson, Laura and Xiao, Tete and Whitehead, Spencer and Berg, Alexander C and Lo, Wan-Yen and others},
  booktitle={Proceedings of the IEEE/CVF international conference on computer vision},
  pages={4015--4026},
  year={2023}
}

@article{lu2024omniparser,
  title={{OmniParser for Pure Vision Based GUI Agent}},
  author={Lu, Yadong and Yang, Jianwei and Shen, Yelong and Awadallah, Ahmed},
  journal={arXiv preprint arXiv:2408.00203},
  year={2024}
}

@article{qin2025ui,
  title={{UI-TARS: Pioneering Automated GUI Interaction with Native Agents}},
  author={Qin, Yujia and Ye, Yining and Fang, Junjie and Wang, Haoming and Liang, Shihao and Tian, Shizuo and Zhang, Junda and Li, Jiahao and Li, Yunxin and Huang, Shijue and others},
  journal={arXiv preprint arXiv:2501.12326},
  year={2025}
}

@article{luo2025gui,
  title={{GUI-R1: A Generalist R1-Style Vision-Language Action Model For GUI Agents}},
  author={Luo, Run and Wang, Lu and He, Wanwei and Chen, Longze and Li, Jiaming and Xia, Xiaobo},
  journal={arXiv preprint arXiv:2504.10458},
  year={2025}
}

@misc{guo2025continuallearninggenerativeai,
      title={{Continual Learning for Generative AI: From LLMs to MLLMs and Beyond}}, 
      author={Haiyang Guo and Fanhu Zeng and Fei Zhu and Jiayi Wang and Xukai Wang and Jingang Zhou and Hongbo Zhao and Wenzhuo Liu and Shijie Ma and Da-Han Wang and Xu-Yao Zhang and Cheng-Lin Liu},
      year={2025},
      eprint={2506.13045},
      archivePrefix={arXiv},
      primaryClass={cs.LG},
      url={https://arxiv.org/abs/2506.13045}, 
}

@article{liu2025llava,
  title={{LLaVA-c: Continual Improved Visual Instruction Tuning}},
  author={Liu, Wenzhuo and Zhu, Fei and Guo, Haiyang and Wei, Longhui and Liu, Cheng-Lin},
  journal={arXiv preprint arXiv:2506.08666},
  year={2025}
}

@article{shenfeld2025rl,
  title={{RL's Razor: Why Online Reinforcement Learning Forgets Less}},
  author={Shenfeld, Idan and Pari, Jyothish and Agrawal, Pulkit},
  journal={arXiv preprint arXiv:2509.04259},
  year={2025}
}

@article{jin2025rl,
  title={{RL Fine-Tuning Heals OOD Forgetting in SFT}},
  author={Jin, Hangzhan and Luan, Sitao and Lyu, Sicheng and Rabusseau, Guillaume and Rabbany, Reihaneh and Precup, Doina and Hamdaqa, Mohammad},
  journal={arXiv preprint arXiv:2509.12235},
  year={2025}
}

@article{liu2025visual,
  title={{Visual-RFT: Visual Reinforcement Fine-Tuning}},
  author={Liu, Ziyu and Sun, Zeyi and Zang, Yuhang and Dong, Xiaoyi and Cao, Yuhang and Duan, Haodong and Lin, Dahua and Wang, Jiaqi},
  journal={arXiv preprint arXiv:2503.01785},
  year={2025}
}

@article{zhang2025reinforcement,
  title={{Why Reinforcement Fine-Tuning Enables MLLMs Preserve Prior Knowledge Better: A Data Perspective}},
  author={Zhang, Zhihao and Dong, Qiaole and Zhang, Qi and Zhao, Jun and Zhou, Enyu and Xi, Zhiheng and Jin, Senjie and Fan, Xiaoran and Zhou, Yuhao and Wu, Mingqi and others},
  journal={arXiv preprint arXiv:2506.23508},
  year={2025}
}

@article{guo2025deepseek,
  title={{DeepSeek-R1: Incentivizing Reasoning Capability in LLMs via Reinforcement Learning}},
  author={Guo, Daya and Yang, Dejian and Zhang, Haowei and Song, Junxiao and Zhang, Ruoyu and Xu, Runxin and Zhu, Qihao and Ma, Shirong and Wang, Peiyi and Bi, Xiao and others},
  journal={arXiv preprint arXiv:2501.12948},
  year={2025}
}

@article{lu2025ui,
  title={{UI-R1: Enhancing Efficient Action Prediction of GUI Agents by Reinforcement Learning}},
  author={Lu, Zhengxi and Chai, Yuxiang and Guo, Yaxuan and Yin, Xi and Liu, Liang and Wang, Hao and Xiao, Han and Ren, Shuai and Xiong, Guanjing and Li, Hongsheng},
  journal={arXiv preprint arXiv:2503.21620},
  year={2025}
}

@article{liu2025infigui,
  title={{InfiGUI-R1: Advancing Multimodal GUI Agents from Reactive Actors to Deliberative Reasoners}},
  author={Liu, Yuhang and Li, Pengxiang and Xie, Congkai and Hu, Xavier and Han, Xiaotian and Zhang, Shengyu and Yang, Hongxia and Wu, Fei},
  journal={arXiv preprint arXiv:2504.14239},
  year={2025}
}

@article{zhou2025gui,
  title={{GUI-G1: Understanding R1-Zero-Like Training for Visual Grounding in GUI Agents}},
  author={Zhou, Yuqi and Dai, Sunhao and Wang, Shuai and Zhou, Kaiwen and Jia, Qinglin and Xu, Jun},
  journal={arXiv preprint arXiv:2505.15810},
  year={2025}
}

@article{bai2025qwen2,
  title={{Qwen2.5-VL Technical Report}},
  author={Bai, Shuai and Chen, Keqin and Liu, Xuejing and Wang, Jialin and Ge, Wenbin and Song, Sibo and Dang, Kai and Wang, Peng and Wang, Shijie and Tang, Jun and others},
  journal={arXiv preprint arXiv:2502.13923},
  year={2025}
}

@article{yuan2025enhancing,
  title={{Enhancing Visual Grounding for GUI Agents via Self-Evolutionary Reinforcement Learning}},
  author={Yuan, Xinbin and Zhang, Jian and Li, Kaixin and Cai, Zhuoxuan and Yao, Lujian and Chen, Jie and Wang, Enguang and Hou, Qibin and Chen, Jinwei and Jiang, Peng-Tao and others},
  journal={arXiv preprint arXiv:2505.12370},
  year={2025}
}

@inproceedings{kapoor2024omniact,
  title={{OmniACT: A Dataset and Benchmark for Enabling Multimodal Generalist Autonomous Agents for Desktop and Web}},
  author={Kapoor, Raghav and Butala, Yash Parag and Russak, Melisa and Koh, Jing Yu and Kamble, Kiran and AlShikh, Waseem and Salakhutdinov, Ruslan},
  booktitle={European Conference on Computer Vision},
  pages={161--178},
  year={2024},
  organization={Springer}
}

@inproceedings{li2025screenspot,
  title={{ScreenSpot-Pro: GUI Grounding for Professional High-Resolution Computer Use}},
  author={Li, Kaixin and Meng, Ziyang and Lin, Hongzhan and Luo, Ziyang and Tian, Yuchen and Ma, Jing and Huang, Zhiyong and Chua, Tat-Seng},
  booktitle={Proceedings of the 33rd ACM International Conference on Multimedia},
  pages={8778--8786},
  year={2025}
}

@article{ye2025gui,
  title={{GUI-ARP: Enhancing Grounding with Adaptive Region Perception for GUI Agents}},
  author={Ye, Xianhang and Li, Yiqing and Dai, Wei and Liu, Miancan and Chen, Ziyuan and Han, Zhangye and Min, Hongbo and Ren, Jinkui and Zhang, Xiantao and Yang, Wen and others},
  journal={arXiv preprint arXiv:2509.15532},
  year={2025}
}

@article{guiaif,
  title={{Continual GUI Agents}},
  author={Liu, Ziwei and Kang, Borui and Yuan, Hangjie and Zhao, Zixiang and Li, Wei and Zhu, Yifan and Feng, Tao},
  year={2026},
  eprint={2601.20732},
  journal={arXiv},
  url={https://arxiv.org/abs/2601.20732}, 
}

@article{yang2025dcpo,
  title={{DCPO: Dynamic Clipping Policy Optimization}},
  author={Yang, Shihui and Dou, Chengfeng and Guo, Peidong and Lu, Kai and Ju, Qiang and Deng, Fei and Xin, Rihui},
  journal={arXiv preprint arXiv:2509.02333},
  year={2025}
}

@article{chen2025empirical,
  title={{An Empirical Study on Eliciting and Improving R1-like Reasoning Models}},
  author={Chen, Zhipeng and Min, Yingqian and Zhang, Beichen and Chen, Jie and Jiang, Jinhao and Cheng, Daixuan and Zhao, Wayne Xin and Liu, Zheng and Miao, Xu and Lu, Yang and others},
  journal={arXiv preprint arXiv:2503.04548},
  year={2025}
}

@article{yu2025dapo,
  title={{DAPO: An Open-Source LLM Reinforcement Learning System at Scale}},
  author={Yu, Qiying and Zhang, Zheng and Zhu, Ruofei and Yuan, Yufeng and Zuo, Xiaochen and Yue, Yu and Dai, Weinan and Fan, Tiantian and Liu, Gaohong and Liu, Lingjun and others},
  journal={arXiv preprint arXiv:2503.14476},
  year={2025}
}

@article{wang2025icpo,
  title={{ICPO: Intrinsic Confidence-Driven Group Relative Preference Optimization for Efficient Reinforcement Learning}},
  author={Wang, Jinpeng and Li, Chao and Ye, Ting and Zhang, Mengyuan and Liu, Wei and Luan, Jian},
  journal={arXiv preprint arXiv:2511.21005},
  year={2025}
}

@article{lian2025ui,
  title={{UI-AGILE: Advancing GUI Agents with Effective Reinforcement Learning and Precise Inference-Time Grounding}},
  author={Lian, Shuquan and Wu, Yuhang and Ma, Jia and Ding, Yifan and Song, Zihan and Chen, Bingqi and Zheng, Xiawu and Li, Hui},
  journal={arXiv preprint arXiv:2507.22025},
  year={2025}
}

\appendix

\newpage
\section{Reward Function and Its Interaction with the Method} 
\label{app:reward} 

Following GUI-AiF~\cite{guiaif}, we employ two complementary continuous rewards to evaluate each rollout prediction with respect to the ground-truth target: a Gaussian Point Reward and a Gaussian Coverage Reward. These rewards provide dense spatial feedback beyond the binary criterion used for GUI grounding evaluation. 

\textbf{Gaussian Representation.} For a bounding box $\mathbf{b}=[x_1,y_1,x_2,y_2]$, we represent the corresponding GUI element as a two-dimensional Gaussian distribution $ \mathcal{N}(\mathbf{x}; \boldsymbol{\mu},\boldsymbol{\Sigma}),$ where $\boldsymbol{\mu} = \left( \frac{x_1+x_2}{2}, \frac{y_1+y_2}{2} \right)$ is the center of the bounding box, and $ \boldsymbol{\Sigma} = \begin{bmatrix} \sigma_x^2 & 0\\ 0 & \sigma_y^2 \end{bmatrix}.$ Following the adaptive variance mechanism of GUI-G$^2$, the standard deviations are determined by the size of the corresponding element: $ \sigma_x = \alpha(x_2-x_1),$ $\sigma_y = \alpha(y_2-y_1), $where we follow GUI-G$^2$ and set $\alpha=0.5$. 

\textbf{Gaussian Point Reward.} For training instance $i$ and rollout $g$, let $\boldsymbol{\mu}^{p}_{i,g} =(c^{p}_{x,i,g},c^{p}_{y,i,g})$ and $\boldsymbol{\mu}^{gt}_{i} =(c^{gt}_{x,i},c^{gt}_{y,i})$ denote the centers of the predicted and ground-truth bounding boxes, respectively. The point-level grounding reward evaluates the predicted center under the Gaussian distribution defined by the ground-truth element: \begin{equation} r^{p}_{i,g} = \exp \left( -\frac{1}{2} \left[ \frac{ (c^{p}_{x,i,g}-c^{gt}_{x,i})^2 }{ (\sigma^{gt}_{x,i})^2 } + \frac{ (c^{p}_{y,i,g}-c^{gt}_{y,i})^2 }{ (\sigma^{gt}_{y,i})^2 } \right] \right). \end{equation} The reward reaches its maximum value when the predicted center coincides with the ground-truth center and decreases smoothly as the prediction moves away from the target. 

\textbf{Gaussian Coverage Reward.} The point reward captures center localization but does not fully characterize the spatial extent of the predicted region. We therefore represent the predicted and ground-truth bounding boxes as $\mathcal{N}^{p}_{i,g} = \mathcal{N} \left( \boldsymbol{\mu}^{p}_{i,g}, \boldsymbol{\Sigma}^{p}_{i,g} \right),$ $\mathcal{N}^{gt}_{i} = \mathcal{N} \left( \boldsymbol{\mu}^{gt}_{i}, \boldsymbol{\Sigma}^{gt}_{i} \right). $ Let $\boldsymbol{\Sigma}^{m}_{i,g} = \frac{ \boldsymbol{\Sigma}^{p}_{i,g} + \boldsymbol{\Sigma}^{gt}_{i} }{2}.$ Following GUI-G$^2$, we measure the overlap between the two Gaussian distributions using the Bhattacharyya coefficient. The resulting coverage reward is \begin{equation} \begin{aligned} r^{r}_{i,g} = \exp\Bigg( &-\frac{1}{8} \left( \boldsymbol{\mu}^{p}_{i,g} - \boldsymbol{\mu}^{gt}_{i} \right)^{T} \left( \boldsymbol{\Sigma}^{m}_{i,g} \right)^{-1} \left( \boldsymbol{\mu}^{p}_{i,g} - \boldsymbol{\mu}^{gt}_{i} \right) \\ &-\frac{1}{2} \ln \frac{ \det(\boldsymbol{\Sigma}^{m}_{i,g}) }{ \sqrt{ \det(\boldsymbol{\Sigma}^{p}_{i,g}) \det(\boldsymbol{\Sigma}^{gt}_{i}) } } \Bigg). \end{aligned} \end{equation} This reward jointly considers center alignment and the spatial extent of the predicted and target regions. 

\textbf{Combined Grounding Reward.} GUI-G$^2$ combines the point and coverage rewards as \begin{equation} r_{i,g} = \nu r^{p}_{i,g} + \gamma r^{r}_{i,g}, \end{equation} where $\nu$ and $\gamma$ control the relative contributions of the two components. Following the default GUI-G$^2$ setting, we use $\nu=\gamma=1.0$, yielding $ r_{i,g} = r^{p}_{i,g} + r^{r}_{i,g}.$ Both reward components lie in $[0,1]$, and therefore $r_{i,g}\in[0,2]$. 

\section{Reward Dynamics Analysis.}
\begin{figure}[h]
  \centering
  \includegraphics[width=\linewidth]{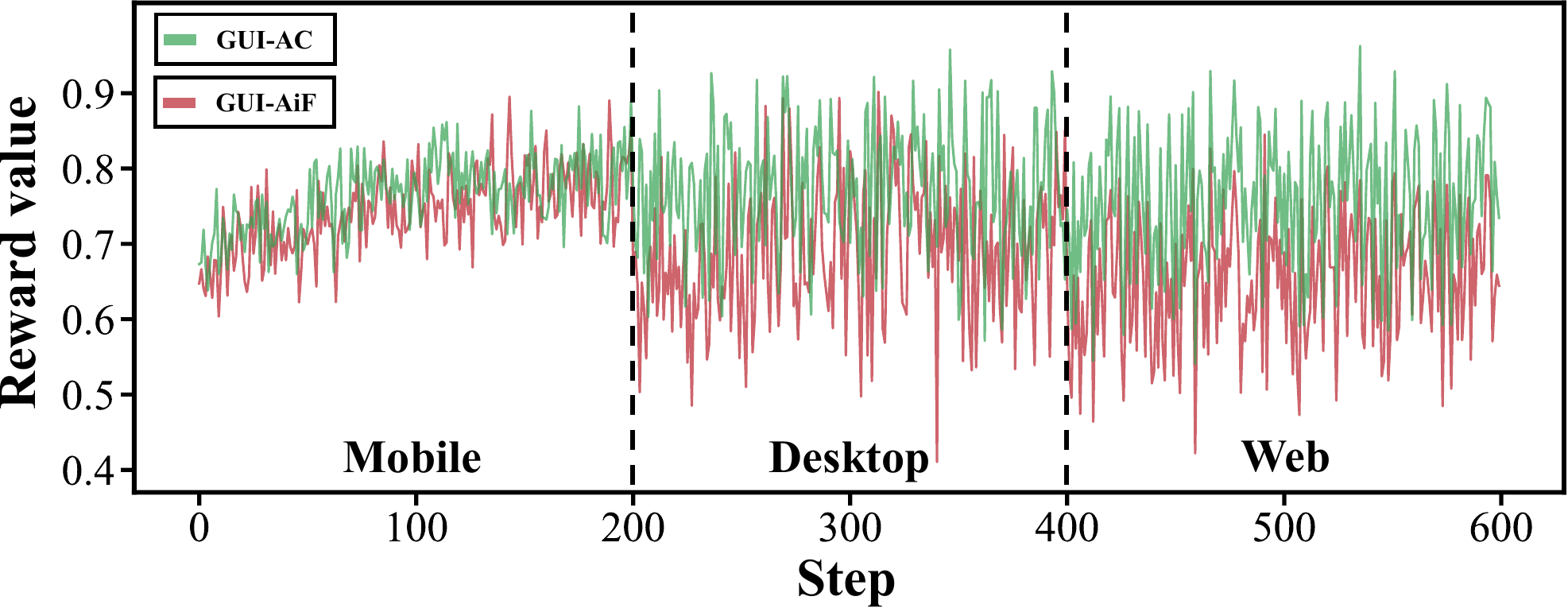}
  \caption{GUI-AC maintains consistently smooth, high-reward trajectories and recovers rapidly after each domain shift, whereas GUI-AiF suffers sharp fluctuations and high variance, highlighting the superior optimization stability of GUI-AC under continual cross-domain learning.}   
  \label{fig:reward}
\end{figure}

We analyze reward dynamics as a proxy for optimization stability in cross-domain continual learning. Figure~\ref{fig:reward} reports per-step rewards over the first 200 training steps of each stage. In the initial Mobile stage, both methods yield relatively steady rewards, consistent with stable on-policy improvement when the interface distribution remains largely stationary. After each domain switch, GUI-AiF exhibits sharp reward discontinuities and high-variance oscillations, producing a jagged trajectory that points to unstable rollout outcomes under distribution shift. In contrast, GUI-AC preserves smooth reward trajectories across all stages: it rapidly re-enters a high-reward regime and sustains a tight reward band with markedly reduced fluctuation amplitude. Taken together, these dynamics indicate that GUI-AC delivers more robust policy optimization in the face of continual distribution shift.

\section{More Experiments}

\subsection{Statistical Validation Across Multiple Seeds} 
\label{app:multiseed}

To evaluate the robustness of the reported improvements, we repeat the primary comparison between GUI-AiF~\cite{guiaif} and GUI-AC using three independent random seeds. Tables~\ref{tab:multiseed} report the mean and standard deviation across the three runs on ScreenSpot-V1 and ScreenSpot-V2, respectively. GUI-AC consistently improves the stage-wise average accuracy over GUI-AiF across both benchmarks.

\begin{table*}[t]
    \centering
    \caption{Continual domain performance (\%) on ScreenSpot-V1 and ScreenSpot-V2. Results are reported as mean $\pm$ standard deviation over three independent runs. M., D., and W. denote the Mobile, Desktop, and Web domain tasks, respectively.}
    \label{tab:multiseed}
    \vspace{-1mm}

    \begin{tabular}{@{}l l cc cc cc c@{}}
        \toprule

        \multirow{2}{*}{\textbf{Method}} &
        \multirow{2}{*}{} &
        \multicolumn{2}{c}{\textbf{Mobile}} &
        \multicolumn{2}{c}{\textbf{Desktop}} &
        \multicolumn{2}{c}{\textbf{Web}} &
        \multirow{2}{*}{\textbf{Avg.}} \\

        \cmidrule(lr){3-4}
        \cmidrule(lr){5-6}
        \cmidrule(lr){7-8}

        & & Text & Icon & Text & Icon & Text & Icon & \\

        \midrule

        \multicolumn{9}{c}{\textbf{ScreenSpot-V1}} \\
        \midrule

        \multirow{3}{*}{GUI-AiF~\cite{guiaif}}
        & M.
        & $92.86{\pm}0.50$
        & $78.57{\pm}0.43$
        & $81.81{\pm}0.49$
        & $68.45{\pm}0.40$
        & $81.46{\pm}0.26$
        & $65.05{\pm}0.44$
        & $78.04{\pm}0.42$ \\

        & M.$\rightarrow$D.
        & $94.05{\pm}0.28$
        & $77.22{\pm}0.28$
        & $93.13{\pm}0.20$
        & $69.26{\pm}0.31$
        & $81.84{\pm}0.41$
        & $66.13{\pm}0.53$
        & $80.27{\pm}0.33$ \\

        & M.$\rightarrow$D.$\rightarrow$W.
        & $96.00{\pm}0.39$
        & $76.75{\pm}0.21$
        & $95.56{\pm}0.28$
        & $68.29{\pm}0.28$
        & $84.79{\pm}0.21$
        & $68.15{\pm}0.43$
        & $81.59{\pm}0.29$ \\

        \midrule

        \multirow{3}{*}{\textbf{GUI-\textcolor[HTML]{C71030}{A}\textcolor[HTML]{3170B5}{C}}}
        & M.
        & $94.21{\pm}0.26$
        & $81.28{\pm}0.55$
        & $86.71{\pm}0.27$
        & $70.02{\pm}0.24$
        & $82.37{\pm}0.24$
        & $70.10{\pm}0.40$
        & $80.78{\pm}0.32$ \\

        & M.$\rightarrow$D.
        & $95.27{\pm}0.31$
        & $81.37{\pm}0.35$
        & $92.27{\pm}0.36$
        & $70.80{\pm}0.45$
        & $87.86{\pm}0.25$
        & $69.17{\pm}0.32$
        & $82.79{\pm}0.34$ \\

        & M.$\rightarrow$D.$\rightarrow$W.
        & $96.52{\pm}0.42$
        & $82.46{\pm}0.59$
        & $94.34{\pm}0.29$
        & $71.53{\pm}0.25$
        & $88.74{\pm}0.49$
        & $71.00{\pm}0.52$
        & $84.10{\pm}0.42$ \\

        \midrule

        \multicolumn{9}{c}{\textbf{ScreenSpot-V2}} \\
        \midrule

        \multirow{3}{*}{GUI-AiF~\cite{guiaif}}
        & M.
        & $93.26{\pm}0.54$
        & $82.09{\pm}0.50$
        & $88.39{\pm}0.30$
        & $73.50{\pm}0.32$
        & $80.22{\pm}0.44$
        & $67.87{\pm}0.63$
        & $80.89{\pm}0.45$ \\

        & M.$\rightarrow$D.
        & $95.83{\pm}0.49$
        & $79.49{\pm}0.38$
        & $89.54{\pm}0.50$
        & $77.01{\pm}0.50$
        & $81.07{\pm}0.26$
        & $68.21{\pm}0.21$
        & $81.86{\pm}0.39$ \\

        & M.$\rightarrow$D.$\rightarrow$W.
        & $96.48{\pm}0.32$
        & $83.60{\pm}0.52$
        & $90.16{\pm}0.41$
        & $77.35{\pm}0.30$
        & $82.81{\pm}0.44$
        & $70.38{\pm}0.29$
        & $83.47{\pm}0.38$ \\

        \midrule

        \multirow{3}{*}{\textbf{GUI-\textcolor[HTML]{C71030}{A}\textcolor[HTML]{3170B5}{C}}}
        & M.
        & $95.25{\pm}0.24$
        & $82.13{\pm}0.41$
        & $89.30{\pm}0.38$
        & $75.10{\pm}0.47$
        & $82.67{\pm}0.30$
        & $68.61{\pm}0.52$
        & $82.18{\pm}0.38$ \\

        & M.$\rightarrow$D.
        & $96.55{\pm}0.42$
        & $82.53{\pm}0.40$
        & $92.39{\pm}0.26$
        & $77.28{\pm}0.39$
        & $83.86{\pm}0.33$
        & $68.64{\pm}0.52$
        & $83.54{\pm}0.38$ \\

        & M.$\rightarrow$D.$\rightarrow$W.
        & $97.73{\pm}0.27$
        & $82.61{\pm}0.20$
        & $95.45{\pm}0.52$
        & $77.25{\pm}0.41$
        & $86.43{\pm}0.54$
        & $71.54{\pm}0.22$
        & $85.17{\pm}0.36$ \\

        \bottomrule
    \end{tabular}
\end{table*}

\subsection{Continual Learning Metrics}
\label{app:continual_metrics}

We evaluate forward transfer (FWT) and backward transfer (BWT) to further analyze the continual learning performance of GUI AC. Let $R_{t,j}$ denote accuracy on task $j$ after stage $t$, and $b_j$ the corresponding accuracy of the common Qwen2.5-VL-3B initialization. Text and icon accuracies are equally weighted within each task. With $T=3$, we define:
$\mathrm{FWT} = \frac{1}{T-1}\sum_{j=2}^{T}(R_{j-1,j}-b_j)$
and
$\mathrm{BWT} = \frac{1}{T-1}\sum_{j=1}^{T-1}(R_{T,j}-R_{j,j}).$
FWT measures transfer to upcoming tasks before training on them, while BWT measures the effect of subsequent training on earlier tasks.

Table~\ref{tab:continual_metrics} shows that GUI-AC achieves higher FWT on both benchmarks while maintaining positive BWT, supporting improved transfer to upcoming tasks and average performance gains on earlier tasks after subsequent training.

\begin{table}[H]
    \centering
    \caption{Continual learning metrics in percentage points,
    computed from the rounded three-run mean accuracies in
    Table~\ref{tab:multiseed}. Higher is better.}
    \label{tab:continual_metrics}
    \setlength{\tabcolsep}{5pt}
    \begin{tabular}{@{}llrr@{}}
        \toprule
        Benchmark & Method & FWT $\uparrow$ & BWT $\uparrow$ \\
        \midrule
        ScreenSpot-V1 & GUI-AiF & 4.43 & 0.70 \\
                     & GUI-AC  & \textbf{8.32} & \textbf{1.57} \\
        \midrule
        ScreenSpot-V2 & GUI-AiF & 8.07 & 1.42 \\
                     & GUI-AC  & \textbf{9.50} & \textbf{1.50} \\
        \bottomrule
    \end{tabular}
\end{table}

\begin{table*}[t]
    \centering
    \caption{Continual domain performance (\%) of different RL optimization variants built upon GUI-AiF on ScreenSpot V1 and ScreenSpot V2. M., D., and W. denote Mobile, Desktop, and Web tasks, respectively.}
    \label{tab:rlbaseline}

    \resizebox{\textwidth}{!}{
    \begin{tabular}{@{}l l cc cc cc c cc cc cc c@{}}
        \toprule

        \multirow{3}{*}{\textbf{Variant}} &
        \multirow{3}{*}{} &
        \multicolumn{7}{c}{\textbf{SSv1 Accuracy (\%)}} &
        \multicolumn{7}{c}{\textbf{SSv2 Accuracy (\%)}} \\

        \cmidrule(lr){3-9}
        \cmidrule(lr){10-16}

        & &
        \multicolumn{2}{c}{Mobile} &
        \multicolumn{2}{c}{Desktop} &
        \multicolumn{2}{c}{Web} &
        \multirow{2}{*}{\textbf{Avg.}} &
        \multicolumn{2}{c}{Mobile} &
        \multicolumn{2}{c}{Desktop} &
        \multicolumn{2}{c}{Web} &
        \multirow{2}{*}{\textbf{Avg.}} \\

        \cmidrule(lr){3-4}
        \cmidrule(lr){5-6}
        \cmidrule(lr){7-8}
        \cmidrule(lr){10-11}
        \cmidrule(lr){12-13}
        \cmidrule(lr){14-15}

        & & Text & Icon & Text & Icon & Text & Icon & &
        Text & Icon & Text & Icon & Text & Icon & \\

        \midrule
        \rowcolor{gray!20}
        \multicolumn{16}{l}{\textit{Standard RL Optimization}} \\

        \multirow{3}{*}{\textbf{$N=8$}}
        & M.
        & 94.5 & 79.5 & 88.1 & 65.0 & 81.6 & 63.6 & 78.7
        & 95.2 & 80.1 & 88.1 & 71.7 & 82.1 & 65.5 & 80.5 \\

        & M.$\rightarrow$D.
        & 95.2 & 80.3 & 89.2 & 65.6 & 82.9 & 65.1 & 79.7
        & 96.6 & 81.0 & 89.6 & 71.0 & 80.3 & 68.0 & 81.1 \\

        & M.$\rightarrow$D.$\rightarrow$W.
        & 95.6 & 79.0 & 92.7 & 66.4 & 84.8 & 68.6 & 81.2
        & 96.9 & 81.9 & 93.8 & 72.4 & 84.2 & 68.3 & 82.9 \\

        \midrule

        \multirow{3}{*}{\textbf{$\eta=5\times10^{-7}$}}
        & M.
        & 94.3 & 78.8 & 87.5 & 64.2 & 80.9 & 62.8 & 78.1
        & 94.8 & 79.5 & 87.3 & 70.8 & 81.4 & 64.8 & 79.8 \\

        & M.$\rightarrow$D.
        & 94.8 & 79.5 & 88.6 & 64.8 & 82.1 & 64.2 & 79.0
        & 96.2 & 80.3 & 88.9 & 70.2 & 79.5 & 67.3 & 80.4 \\

        & M.$\rightarrow$D.$\rightarrow$W.
        & 95.2 & 78.2 & 91.9 & 65.6 & 84.0 & 67.5 & 80.4
        & 96.5 & 81.0 & 92.9 & 71.4 & 83.4 & 67.5 & 82.1 \\

        \midrule

        \multirow{3}{*}{\textbf{$\epsilon=0.4$}}
        & M.
        & 94.0 & 78.0 & 86.8 & 63.5 & 80.0 & 61.8 & 77.4
        & 94.3 & 78.8 & 86.4 & 70.0 & 80.5 & 63.8 & 79.0 \\

        & M.$\rightarrow$D.
        & 94.5 & 78.8 & 87.9 & 64.0 & 81.2 & 63.2 & 78.3
        & 95.8 & 79.5 & 88.1 & 69.4 & 78.8 & 66.5 & 79.7 \\

        & M.$\rightarrow$D.$\rightarrow$W.
        & 94.9 & 77.5 & 91.2 & 64.8 & 83.2 & 66.4 & 79.7
        & 96.1 & 80.1 & 92.1 & 70.4 & 82.5 & 66.8 & 81.3 \\

        \midrule
        \midrule
        \rowcolor{gray!20}
        \multicolumn{16}{l}{\textit{General RL Optimization}} \\

        \multirow{3}{*}{DAPO~\cite{yu2025dapo}}
        & M.
        & 93.0 & 78.0 & 81.5 & 67.8 & 81.0 & 65.5 & 77.8
        & 93.2 & 81.5 & 87.8 & 72.8 & 79.8 & 66.1 & 80.2 \\

        & M.$\rightarrow$D.
        & 94.0 & 77.5 & 92.0 & 68.5 & 81.5 & 65.3 & 79.8
        & 95.4 & 79.0 & 89.0 & 76.2 & 80.7 & 69.3 & 81.6 \\

        & M.$\rightarrow$D.$\rightarrow$W.
        & 95.8 & 77.2 & 93.2 & 68.7 & 84.1 & 67.4 & 81.1
        & 96.4 & 82.8 & 90.8 & 77.0 & 82.7 & 68.9 & 83.1 \\

        \midrule

        \multirow{3}{*}{DCPO~\cite{yang2025dcpo}}
        & M.
        & 93.5 & 79.2 & 83.0 & 68.8 & 81.8 & 65.9 & 78.7
        & 94.0 & 82.0 & 88.8 & 73.8 & 80.5 & 66.9 & 81.0 \\

        & M.$\rightarrow$D.
        & 94.8 & 79.0 & 92.5 & 69.4 & 83.5 & 64.4 & 80.6
        & 96.0 & 80.5 & 91.0 & 77.0 & 81.8 & 67.5 & 82.3 \\

        & M.$\rightarrow$D.$\rightarrow$W.
        & 96.2 & 79.0 & 93.8 & 70.0 & 86.0 & 68.4 & 82.2
        & 97.0 & 83.0 & 92.3 & 77.1 & 83.4 & 68.6 & 83.6 \\

        \midrule

        \multirow{3}{*}{ICPO~\cite{wang2025icpo}}
        & M.
        & 93.2 & 78.8 & 82.2 & 68.0 & 81.0 & 66.0 & 78.2
        & 93.7 & 81.8 & 88.2 & 73.0 & 80.0 & 66.9 & 80.6 \\

        & M.$\rightarrow$D.
        & 94.4 & 78.2 & 92.8 & 68.8 & 82.0 & 64.4 & 80.1
        & 95.8 & 80.0 & 90.5 & 76.8 & 81.2 & 67.1 & 81.9 \\

        & M.$\rightarrow$D.$\rightarrow$W.
        & 96.1 & 78.0 & 94.0 & 69.3 & 85.5 & 66.6 & 81.6
        & 96.8 & 83.2 & 91.8 & 77.4 & 83.8 & 67.4 & 83.4 \\
        
        \midrule
        \bottomrule
    \end{tabular}
    }
\end{table*}

\subsection{Additional RL Baselines.}
\label{app:rlbaseline}

To examine whether the improvements of GUI-AC can be reproduced by generic optimization strategies, we construct a series of optimization variants on top of GUI-AiF~\cite{guiaif}. As shown in Table~\ref{tab:rlbaseline}, we consider two categories of alternatives. The first category, termed \textit{Standard RL Optimization}, modifies commonly used optimization configurations, including a larger rollout group size ($N=8$), a lower learning rate ($\eta=5\times10^{-7}$), and a wider policy clipping margin ($\epsilon=0.4$). The second category, termed \textit{General RL Optimization}, incorporates more advanced RL optimization methods, including DAPO~\cite{yu2025dapo}, DCPO~\cite{yang2025dcpo}, and ICPO~\cite{wang2025icpo}.

\textbf{Standard RL Optimization.} Adjusting standard optimization configurations provides only limited and inconsistent improvements. Increasing the rollout group size can improve several metrics, whereas reducing the learning rate or widening the policy clipping range yields gains only on certain domains or interaction types. However, none of these variants consistently improves performance throughout the continual sequence. This suggests that simply increasing sampling diversity or globally changing the optimization step size and clipping range is insufficient to address the instability introduced by continual GUI distribution shifts.


\textbf{General RL Optimization.}DAPO, DCPO, and ICPO improve several individual metrics over the original GUI-AiF optimization, confirming that generic RL optimization can partially alleviate training instability. However, these improvements are not consistent across domains and interaction types, and even the strongest baseline remains below GUI-AC in final average accuracy. This suggests that the gains of GUI-AC cannot be reproduced solely by generic exploration or confidence-calibration strategies. DAPO promotes low-probability exploration through asymmetric clipping, DCPO adapts clipping according to token-level prior probabilities, and ICPO exploits generation likelihood as intrinsic confidence. These mechanisms do not explicitly distinguish the spatial semantics of GUI grounding outcomes: a low-probability prediction may be either a rare correct localization or an incorrect coordinate, while a high-probability prediction may represent either a stable hit or an outdated spatial pattern. GUI-AC therefore jointly considers grounding advantage and rollout probability to identify failure modes and selectively calibrate their optimization. 

\begin{table*}[h]
    \centering
    \caption{Continual domain performance (\%) under the reversed training sequence on ScreenSpot-V1 and ScreenSpot-V2. W., D., and M. denote Web, Desktop, and Mobile training stages, respectively.}
    \label{tab:reverse_order}
    \vspace{-1mm}
    \resizebox{\textwidth}{!}{
    \begin{tabular}{@{}l l cc cc cc c cc cc cc c@{}}
        \toprule

        \multirow{3}{*}{\textbf{Method}} & \multirow{3}{*}{} &
        \multicolumn{7}{c}{\textbf{SSv1 Accuracy (\%)}} &
        \multicolumn{7}{c}{\textbf{SSv2 Accuracy (\%)}} \\

        \cmidrule(lr){3-9} \cmidrule(lr){10-16}

        & &
        \multicolumn{2}{c}{Mobile} &
        \multicolumn{2}{c}{Desktop} &
        \multicolumn{2}{c}{Web} &
        \multirow{2}{*}{\textbf{Avg.}} &
        \multicolumn{2}{c}{Mobile} &
        \multicolumn{2}{c}{Desktop} &
        \multicolumn{2}{c}{Web} &
        \multirow{2}{*}{\textbf{Avg.}} \\

        \cmidrule(lr){3-4}
        \cmidrule(lr){5-6}
        \cmidrule(lr){7-8}
        \cmidrule(lr){10-11}
        \cmidrule(lr){12-13}
        \cmidrule(lr){14-15}

        & & Text & Icon & Text & Icon & Text & Icon & &
        Text & Icon & Text & Icon & Text & Icon & \\

        \midrule

        \multirow{3}{*}{GUI-AiF~\cite{guiaif}}
        & W.
        & 94.9 & 74.7 & 82.0 & 67.9 & 81.3 & 64.1 & 77.5
        & 95.2 & 77.3 & 82.5 & 72.7 & 81.2 & 64.0 & 78.8 \\

        & W.$\rightarrow$D.
        & 95.6 & 69.4 & 87.6 & 67.1 & 83.7 & 64.8 & 78.0
        & 95.9 & 72.5 & 88.7 & 76.7 & 77.8 & 69.6 & 80.2 \\

        & W.$\rightarrow$D.$\rightarrow$M.
        & 89.7 & 72.6 & 92.6 & 69.6 & 81.4 & 69.5 & 79.2
        & 89.7 & 74.9 & 92.7 & 72.7 & 86.8 & 72.2 & 81.5 \\

        \midrule
        \midrule

        \multirow{3}{*}{\textbf{GUI-\textcolor[HTML]{C71030}{A}\textcolor[HTML]{3170B5}{C}}}
        & W.
        & 96.0 & 77.2 & 84.5 & 69.9
        & 84.3 & 67.4 & 79.9
        & 95.9 & 78.8 & 85.7 & 73.9
        & 83.6 & 65.1 & 80.5 \\
        
        & W.$\rightarrow$D.
        & 96.4 & 72.4 & 90.1 & 69.1
        & 86.9 & 67.7 & 80.4
        & 96.2 & 74.5 & 90.5 & 77.7
        & 79.8 & 71.0 & 81.6 \\
        
        & W.$\rightarrow$D.$\rightarrow$M.
        & \textbf{92.9} & \textbf{75.6} & \textbf{94.6} & \textbf{72.1}
        & \textbf{84.4} & \textbf{71.6} & \textbf{81.9}
        & \textbf{93.8} & \textbf{76.7} & \textbf{95.6} & \textbf{75.9}
        & \textbf{89.1} & \textbf{73.3} & \textbf{84.1} \\
        
        \midrule
        \bottomrule
    \end{tabular}
    }
\end{table*}

\subsection{Robustness to Training Order.}
\label{app:reverse_order}

To examine whether the effectiveness of GUI-AC depends on a specific task order, we conduct an additional experiment by reversing the cross-domain training sequence. As shown in Table~\ref{tab:reverse_order}, GUI-AC consistently achieves higher average accuracy than GUI-AiF at every stage of the reversed training sequence. These results demonstrate that the effectiveness of GUI-AC is not tied to the specific training order used in the main experiments. 

\section{Limitations.}
Although we achieved satisfactory performance on all benchmarks, our research has several limitations. First, in extreme reinforcement-learning settings where the base model initially fails on a new interface, grounding certainty may stay near zero for long periods. In this case, both modules become close to constant adjustments, which limits the benefit of optimization. Second, due to computational resource constraints, we primarily study a 3B-scale model with 3 training datasets and 3 evaluation benchmarks. In future studies, we plan to investigate more complex continual GUI agent scenarios and conduct experiments with larger‑scale base models.

\end{document}